\documentclass[final,5p,times,twocolumn]{elsarticle}

\usepackage{amssymb}
\usepackage{amsthm}
\usepackage{amsmath}
\usepackage{float}
\usepackage{xcolor}
\definecolor{linkcolor}{rgb}{0, 0.498, 0.675}

\usepackage{subfig}
\usepackage{graphicx}
\usepackage{appendix}

\usepackage{multirow}
\usepackage{colortbl}
\usepackage{booktabs}
\newcommand{\CC}[1]{\cellcolor{#1}}
\definecolor{ours}{rgb}{0.98, 0.85, 0.87}
\usepackage{hyperref}
\hypersetup{colorlinks=true, allcolors=red}
\usepackage{cleveref}
\usepackage{bm}
\Crefname{figure}{Fig.}{Figs.}

\newcommand{\FrameworkName}{LoRAC}
\newcommand{\FrameworkNameWo}{\FrameworkName{} w/o TII}
\newcommand{\FrameworkNameIPCWo}{\FrameworkName{}-IPC w/o TII}
\newcommand{\FrameworkNameIPC}{\FrameworkName{}-IPC}

\newcommand{\ie}{i.e.,}
\usepackage{algorithm}
\usepackage{algorithmic}

\usepackage{lineno}

\journal{Pattern Recognition}

\begin{document}

\hypersetup{allcolors=linkcolor}

\begin{frontmatter}



\title{LoRA-Based Continual Learning with Constraints on Critical Parameter Changes}


\author[label1]{Shimou Ling}
\author[label1]{Liang Zhang}
\author[label1]{Jiangwei Zhao}
\author[label1]{Lili Pan\texorpdfstring{\corref{cor1}}{}}
\cortext[cor1]{Corresponding author.}
\ead{lilipan@uestc.edu.cn}
\author[label1]{Hongliang Li}

\affiliation[label1]{organization={
University of Electronic Science and Technology of China},
            city={Chengdu},
            country={China}}

\begin{abstract}
    LoRA-based continual learning represents a promising avenue for leveraging pre-trained models in downstream continual learning tasks. Recent studies have shown that orthogonal LoRA tuning effectively mitigates forgetting. However, this work unveils that under orthogonal LoRA tuning, the critical parameters for pre-tasks still change notably after learning post-tasks.
    To address this problem, we directly propose freezing the most critical parameter matrices in the Vision Transformer (ViT) for pre-tasks before learning post-tasks. 
    In addition, building on orthogonal LoRA tuning, we propose orthogonal LoRA composition (LoRAC) based on QR decomposition, which may further enhance the plasticity of our method. 
    Elaborate ablation studies and extensive comparisons demonstrate the effectiveness of our proposed method. 
    Our results indicate that our method achieves state-of-the-art (SOTA) performance on several well-known continual learning benchmarks. For instance, on the Split CIFAR-100 dataset, our method shows a 6.35\% improvement in accuracy and a 3.24\% reduction in forgetting compared to previous methods. Our code is available at \url{https://github.com/learninginvision/LoRAC-IPC}.

\end{abstract}



\begin{keyword}


Continual learning, Pre-trained model, Orthogonal LoRA composition, Important parameter
\end{keyword}

\end{frontmatter}


\section{Introduction}
\label{sec:intro}
Continual learning (CL) is the process of sequentially training a model on multiple tasks while retaining knowledge acquired from previous tasks~\citep{thrun1995lifelong, chen2018lifelong}. Neural networks often forget knowledge learned from previous tasks after acquiring new knowledge, a phenomenon known as catastrophic forgetting~\citep{mccloskey1989catastrophic}. Significant efforts have been made to alleviate catastrophic forgetting in neural networks in recent years. These studies can be categorized into three main approaches: architecture-based~\citep{rusu2016progressive, fernando2017pathnet, dcpoc, kanets}, regularization-based~\citep{zenke2017continual,ewc,hard, dfd}, and replay-based~\citep{rolnick2019experience,riemer2018learning,buzzega2020dark, rnks}. Despite their proven effectiveness, these approaches still fall short of practical requirements.

\begin{figure}[t!]

    \centering
    \subfloat[The vector product between the columns of $\tilde{\mathbf{Q}}_{T}$.]{\includegraphics[width=0.24\textwidth, page=1]{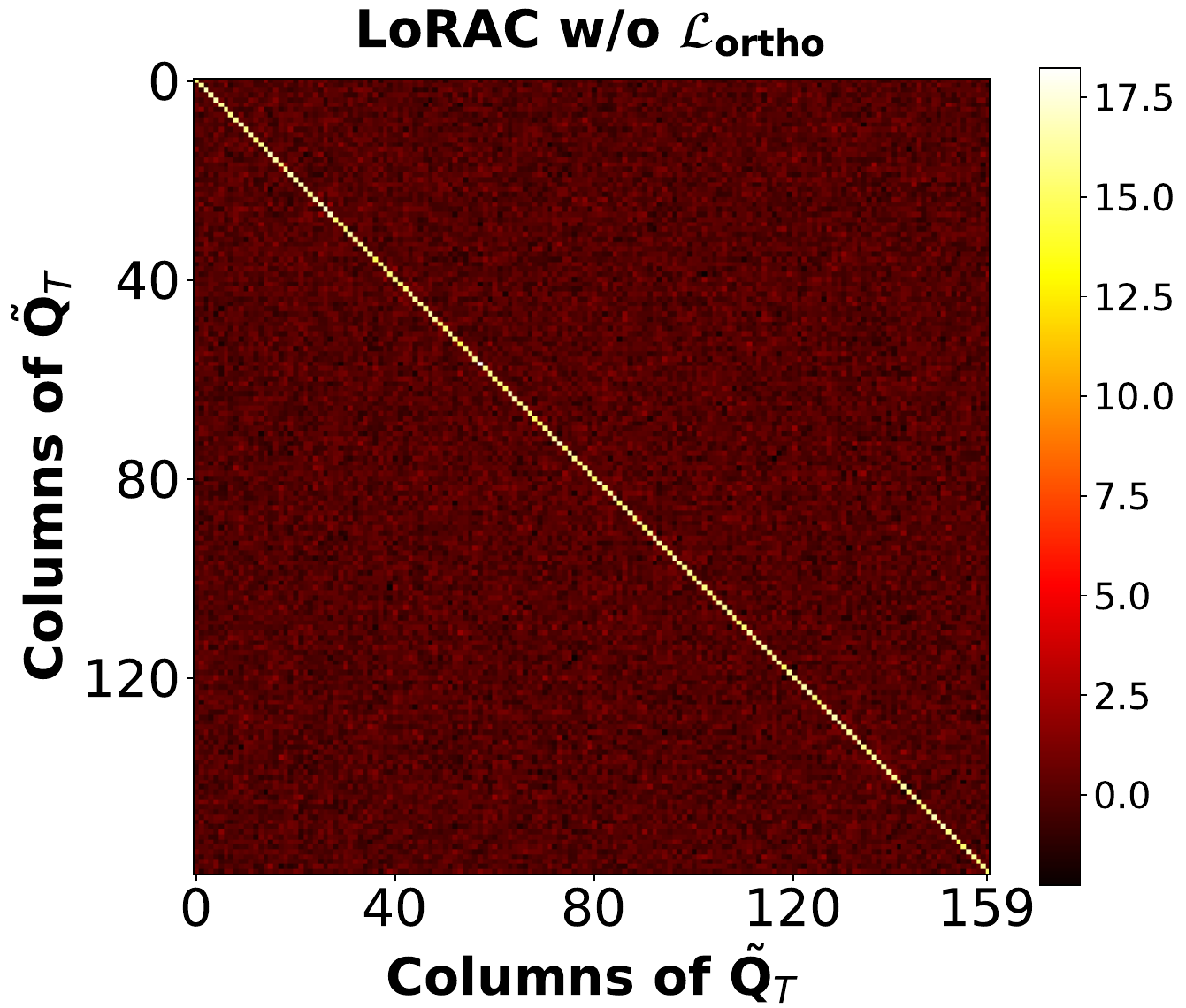}
    
    \includegraphics[width=0.24\textwidth, page=2]{Figures/lorac_hotmap.pdf}}
    \vspace{-5pt}

    \subfloat[The average importance and variation of parameters within each parameter matrix across different blocks.]{\includegraphics[width=0.48\textwidth]{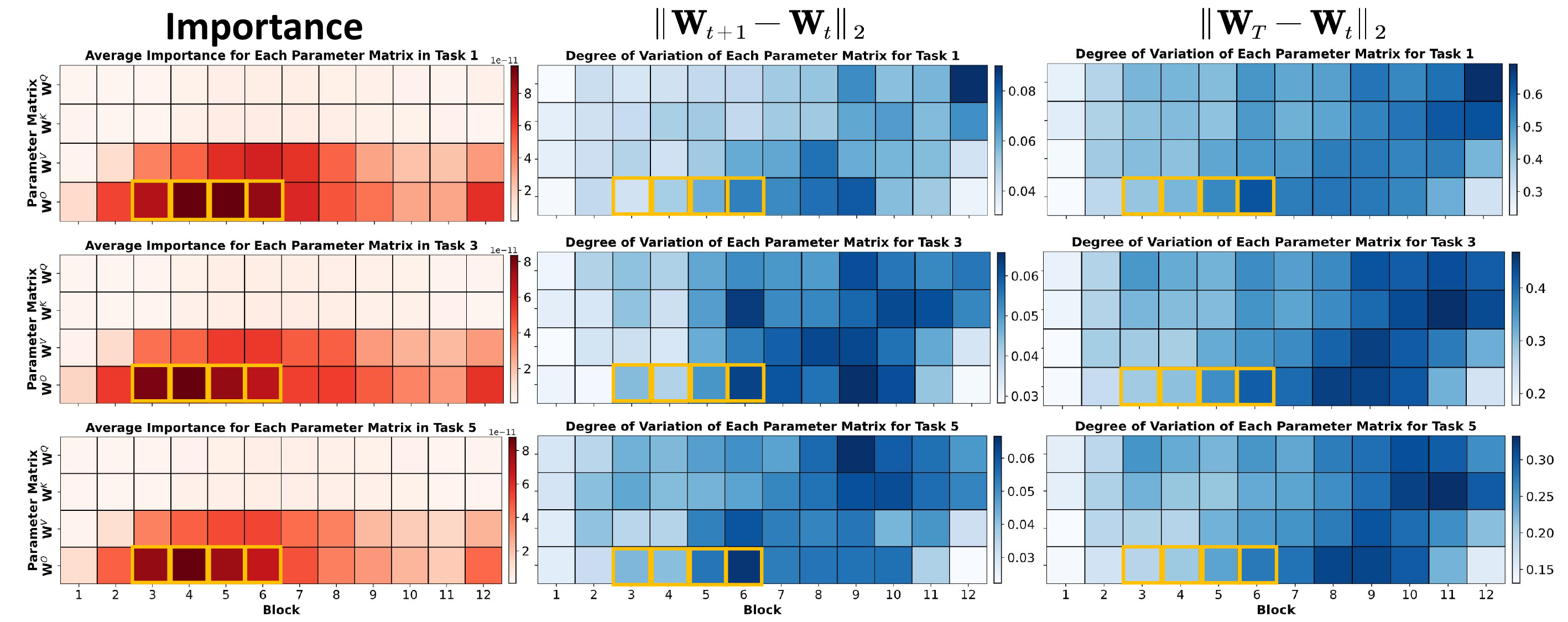}}
    \caption{The degree of variation in important parameters with orthogonal constraints. 
    The calculation of parameter importance is based on the sensitivity of the parameter to training losses and is discussed in \cref{sec:ipc}.
    We use the L2-norm to measure the degree of variation in the parameters after completing task  $t+1$  ($||\mathbf{W}_{t+1} - \mathbf{W}_{t}||_{2}$) and after completing all tasks ($||\mathbf{W}_{T} - \mathbf{W}_{t}||_{2}$), respectively. Important parameters for each task are highlighted by yellow boxes. }
     
    \label{fig:figure1}
    
\vspace{-10pt}
\end{figure}

Over the past year, visual continual learning combined with pre-trained models (PTMs) has demonstrated significant superiority in alleviating forgetting. Prompt tuning has become the most common method to integrate PTMs with continual learning. Early works in this area, including L2P~\citep{wang2022learning}, Dual-Prompt~\citep{wang2022dualprompt}, and S-Prompt~\citep{wang2022sprompts}, initiate the exploration of rehearsal-free continual learning. Later, CODA-Prompt~\citep{smith2023coda} and HiDe-Prompt~\citep{wang2023hide} further advance rehearsal-free continual learning by improving prompt tuning integration. Additionally, LAE~\citep{lae} combines several mainstream tuning methods, such as Adapter~\citep{adapter}, LoRA~\citep{lora}, and Prompt~\citep{prompttuning, prefixtuning}, to explore more efficient tuning methods to alleviate forgetting.

In these methods, LoRA fine-tuning is the most promising due to its cost-efficiency and high-quality tuning results. Recently, a small number of studies~\citep{o-lora, inflora} on orthogonal LoRA tuning have demonstrated their effectiveness in mitigating forgetting. These studies, inspired by orthogonal gradient descent (OGD)~\citep{ogd}, assume that incorporating orthogonal LoRA modules into pre-trained parameters will not alter the training loss of previous tasks. As a result, since the previous training loss remains unchanged during the continual learning process, forgetting can be effectively alleviated.

However, in this work, we find that parameters sensitive to the training loss of previous tasks still change significantly under orthogonal LoRA tuning. This means that the training loss for previous tasks still changes notably, and forgetting is not fully alleviated.


To investigate the root cause, we estimate the average importance of each parameter matrix in the Vision Transformer (ViT) and evaluate their variations across the pre- and post-tasks. 
As depicted in Fig.~\ref{fig:figure1} (b), the parameter matrices important to pre-tasks, highlighted by yellow boxes, still change significantly in continual learning.
On the other hand, we perform QR decomposition of the projection-down matrices of the LoRA modules learned on each task. 
The result in Fig.~\ref{fig:figure1} (a) demonstrates that the low-rank matrices learned on different tasks are orthogonal to each other.
Then a question arises: Why important parameters change significantly under orthogonal LoRA fine-tuning?

We speculate when using LoRA matrices to make a low-rank approximation to the parameter matrices, by setting the rank  much smaller than the original rank, the parameter space corresponding to LoRA fails to represent that of the original parameter matrices. Thus, even if we constrain the LoRA matrices to be orthogonal and obtain a low-rank solution, there is no guarantee that the original parameter matrices are orthogonal.

\begin{figure*}[t!]
    \centering
    \includegraphics[scale=0.42]{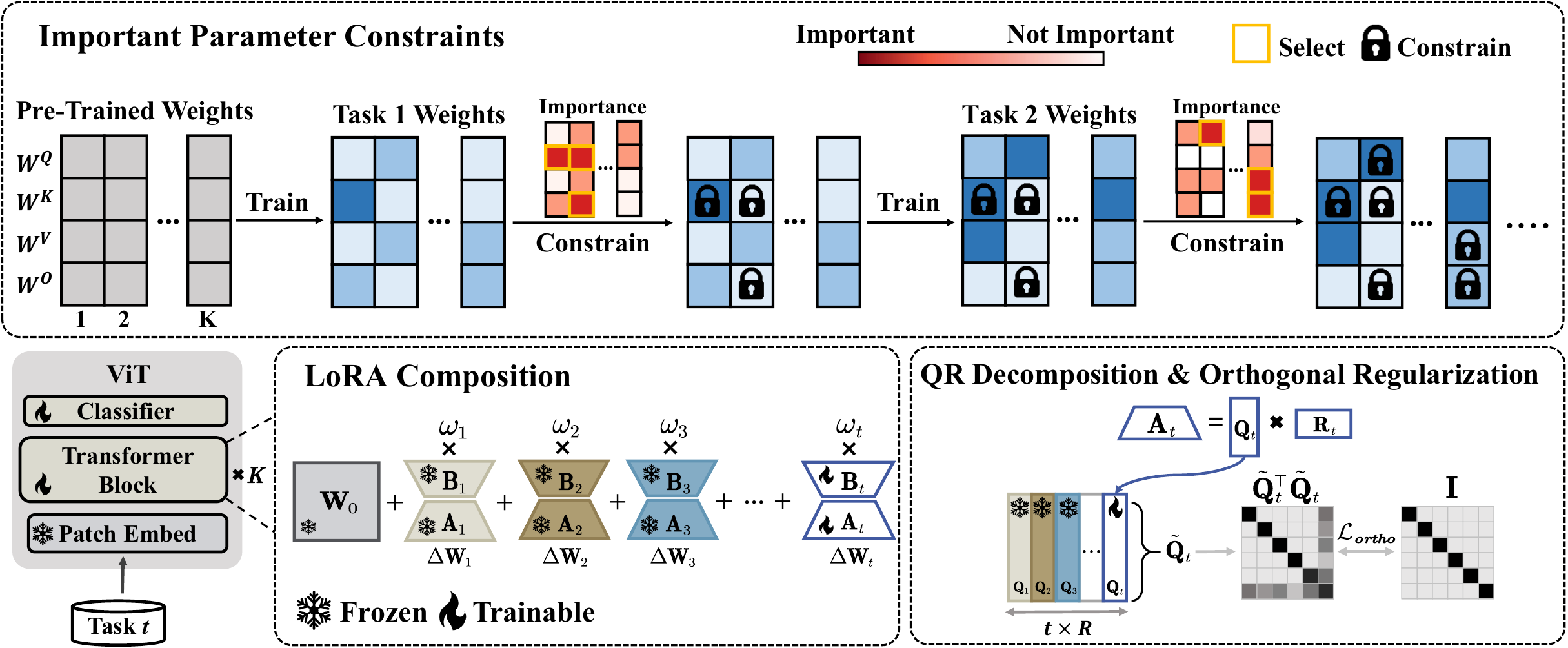}
    \caption{\textbf{Continual Learning with Orthogonal LoRA Composition and Important Parameter Constraints}. \textbf{The upper} illustrates the workflow of Important Parameter Constraints (IPC). Upon completion of training for the current task, parameter matrices important to the current task are constrained to remain unchanged in continual learning. \textbf{The lower} shows the framework for Orthogonal LoRA Composition, consisting of three components: LoRA composition, the QR decomposition of matrix $\mathbf{A}_{t}$ and the orthogonality regularization on the matrix $\tilde{\mathbf{Q}}_t$. }
    \label{fig:lorac-ipc}
    \vspace{-10pt}
\end{figure*}

Based on our analysis, we propose a novel orthogonal LoRA composition method, named \FrameworkNameIPC, for continual learning, which incorporates important parameter change constraints. Fig.~\ref{fig:lorac-ipc} provides an overview of \FrameworkNameIPC. This method integrates parameters from the pre-trained model (PTM) with sequentially learned orthogonal LoRA modules to preserve previous knowledge and incorporate new knowledge. Crucially, constraints are applied to critical ViT parameters to minimize forgetting. By assigning different weights to sequentially learned LoRA modules, we enhance the adaptability of orthogonal LoRA composition. In summary, this work makes the following contributions:

\begin{itemize}
    \item We propose a new PTM-based continual learning method, namely \FrameworkName{}. It explores using orthogonal LoRA composition to preserve previous knowledge and incorporate new knowledge in continual learning. Additionally, by assigning different weights to various LoRA modules, we enhance the adaptability of orthogonal LoRA tuning.
    
    \item  We unveil that in existing continual learning with orthogonal LoRA tuning, the critical parameters sensitive to the current training loss undergo substantial changes across tasks. Based on this finding, we introduce Important Parameter Constraints (IPC) within the LoRAC framework, enhancing the adaptability of orthogonal LoRA tuning.
    \item We conduct detailed ablation studies and comprehensive comparisons to illustrate the effectiveness of \FrameworkNameIPC~across multiple well-established continual learning benchmarks. The results indicate that on Split CIFAR-100, \FrameworkNameIPC{} achieves 6.35\% higher accuracy and reduces forgetting by 3.24\% compared to the previous method when utilizing the widely-used pre-trained model, Sup-21K.
\end{itemize}

\section{Related Work}
\label{sec:rela}

\noindent\textbf{Continual learning}. The objective of continual learning is to enable deep neural networks to continually acquire, update, and accumulate new knowledge, akin to human learning~\citep{wang2024comprehensive}. Nonetheless, existing models often face the \emph{stability–plasticity dilemma}, leading to the issue of \emph{catastrophic forgetting}~\citep{mccloskey1989catastrophic}. Typically, existing algorithms designed to address the aforementioned issue can be classified into three categories~\citep{parisi2019continual}. 
Regularization-based methods~\citep{zenke2017continual,ewc, hard, dfd} are characterized by adding explicit regularization terms that depend on weights or gradients of the previous model to balance the old and new tasks. HARD~\citep{hard} proposes a relaxed distillation constraint in the super-feature space that enhances the model's ability to learn new knowledge while maintaining knowledge of old tasks. DFD~\citep{dfd} approximates the knowledge distillation term using Taylor expansion and implements it as a novel regularizer to penalize parameter changes across training tasks.
Architecture-based methods~\citep{rusu2016progressive, fernando2017pathnet, dcpoc, kanets} dynamically expand the network or isolate specific model parameters that are crucial for different tasks. DCPOC~\citep{dcpoc} utilizes variational autoencoders as feature encoders, enhancing the discriminability of the output from each branch corresponding to each task. KANets~\citep{kanets} freezes the previously trained model to retain existing knowledge and proposes a consistent trainable network as the other branch to learn new concepts.
Rehearsal-based methods~\citep{rolnick2019experience,riemer2018learning,buzzega2020dark, rnks} mitigate catastrophic forgetting by either setting up a memory buffer to store and replay past experience during the learning process of the current task or generating fake samples with an additional generator. 
Experience Replay (ER)~\citep{riemer2018learning} utilizes the reservoir sampling strategy to update the memory buffer and trains the model with the current data and a mini-batch comprised of randomly selected old samples from the memory buffer. DER~\citep{buzzega2020dark} can be viewed as an enhanced iteration of ER, which combines rehearsal and distillation loss to retrain past experiences. 
RNKS~\citep{rnks} retains the model performance on old classes by solving class imbalance problem between stored old exemplars and new training examples in feature space.
Despite their conceptual simplicity, rehearsal-based methods consistently achieve state-of-the-art performance across a variety of benchmarks.

\noindent\textbf{Parameter-Efficient Tuning}.
As an efficient alternative to full fine-tuning, parameter-efficient tuning methods aim to tune the pre-trained models (PTMs) by adjusting lightweight trainable parameters while keeping most pre-trained parameters frozen. Current research endeavors employ diverse methods for introducing lightweight trainable parameters. Adapter~\citep{adapter} is first proposed to insert small newly initialized parameter modules to each transformer layer. 
Prompt tuning~\citep{prompttuning} and Prefix tuning~\citep{prefixtuning} introduce additional trainable prefix tokens, namely prompt, to extend the input of each transformer layer, and tune only the prompts. 
LoRA~\citep{lora} assumes that parameter changes occur within a low-rank space, facilitating the fine-tuning of pre-trained models for downstream tasks by decomposing incremental updates into the multiplication of two low-rank matrices. 
Recent research indicates that LoRA and its variations~\citep{lora-fa, adalora} demonstrate efficiency in parameters and inference, thereby enabling effective fine-tuning of pre-trained models for adaptation to downstream tasks. Our study leverages these advantages by introducing a novel composition approach to implement LoRA in continual learning.

\noindent \textbf{Continual Learning with PTMs}. 
Recent advances in pre-training have made pre-trained models (PTMs) readily available for continual learning.
L2P~\citep{wang2022learning}, Dual-Prompt~\citep{wang2022dualprompt} and CODA-Prompt~\citep{smith2023coda} apply visual prompt tuning~\citep{vpt} to class-incremental learning (CIL) based on the pre-trained Vision Transformer, and the distinction among these three methods is how the prompt is selected and integrated. 
HiDe-Prompt~\citep{wang2023hide} deconstructs the continual learning process into hierarchical components, and in the testing stage, it initially infers the task ID and then makes predictions by using task-specific prompts. CPP~\citep{cpp} optimizes task-specific prompts by introducing a contrastive loss with class prototypes and retrieves task-specific prompts for samples based on the prototypes.
OVOR~\citep{OVOR} proposes virtual outlier regularization to tighten the classifier's decision boundary, thereby alleviating class confusion among different tasks in a rehearsal-free CIL setting.
PGP~\citep{dualp} combines Prompt-tuning with gradient projection to prevent forgetting by reaching the orthogonality condition for the prompt gradient.
CPrompt~\citep{cpompt} aims to bridge the gap between the training and testing stages to enhance prediction robustness and improve prompt selection accuracy.
ConvPrompt~\citep{convprompt} generates task-specific prompts by convolution over task-shared parameters and leverages Large Language Models to dynamically decide the number of prompts to be learned. 

In addition, LAE~\citep{lae} introduces a unified CL framework that unifies several widely used parameter-efficient tuning methods, including Adapter, LoRA, and Prompt. 
ADAM~\citep{Adam} aggregates the embeddings of PTM and adapts models for classifier construction, while SLCA~\citep{zhang2023slca} finds that the model performs better if the learning rate for fine-tuning the ViT backbone is lower than the learning rate for training the classification head.
EASE~\citep{ease} trains a distinct lightweight adapter module for each new task and designs a semantic mapping to complement the drift of old class prototypes.
RanPAC~\citep{RanPAC} proposes a training-free Random Projection layer with nonlinear activation between the pre-trained model’s feature representations and output head, which enhances the linear separability of class features for class-prototype-based CL, thereby effectively mitigating catastrophic forgetting.

The recent proposed work, O-LoRA~\citep{o-lora}, has focused on how to use orthogonal LoRA tuning for continual learning in language models (LMs). However, this work does not provide an effective way for LoRA composition. Besides, InfLoRA~\citep{inflora} has proposed to eliminate the interference from new tasks by ensuring the LoRA of the new task is orthogonal to the inputs of the old task and freezing dimensionality reduction matrices during training.
Our work maintains orthogonal LoRA tuning for continual learning. 

Furthermore, we unveil that even with orthogonal LoRA tuning, the parameters sensitive to the training loss of pre-tasks change significantly in continual learning. This is the first work to identify and investigate this phenomenon. More importantly, we propose an efficient solution to this problem by implementing important parameter constraints.

\section{Continual Learning with Orthogonal LoRA Compostion and Important Parameter Constraints}
\label{sec:Method}

We split the CL classification problem into $T$ tasks $\left\{ \mathcal{T}_{1},\mathcal{T}_2,...,\mathcal{T}_T \right\}$, where each task $\mathcal{T}_t$ is associated with a dataset $\mathcal{D}_t=\{(\mathbf{x}_{t,i},y_{t,i})_{i=1}^{N_t}\}$ containing $N_t$ samples. $\mathbf{x}_{t,i}$ represents the input sample and its corresponding label is denoted as $y_{t,i}$.
Each data pair $(\mathbf{x}_{t,i},y_{t,i})\in (\mathcal{X}_t\times\mathcal{Y}_t)$ belongs to a distribution $(\mathcal{X}_t\times\mathcal{Y}_t)$.
Generally, a neural network model trained on task $t$ can be denoted as an embedding function $f\left(\cdot,\mathbf{\Theta}_t\right) :\mathbb{R}^{W\times H\times C}\rightarrow \mathbb{R}^D$ parameterized by $\mathbf{\Theta}_t$, and a classifier $h\left(\cdot, \boldsymbol{\Phi}_t\right) : \mathbb{R}^{D}\rightarrow \mathbb{R}^{M}$ parameterized by $\boldsymbol{\Phi}_t$, where $D$ represents the feature dimension and $M$ represents the number of classes in each task. The overall objective of continual learning is to train a model $f\left(\cdot, \mathbf{\Theta}_t \right)$ and a classifier $h\left(\cdot, \boldsymbol{\Phi }_t\right)$, capable of predicting labels for an unseen test sample $\mathbf{x}$ from arbitrary tasks seen so far. The data from previous tasks may no longer be available for training in future tasks.

\subsection{Orthogonal LoRA Composition}
\label{sec:ortho loss}


Our approach involves utilizing a pre-trained ViT model as the base model and employing LoRA composition to alleviate catastrophic forgetting. 
The lower part of Fig.~\ref{fig:lorac-ipc} illustrates the framework for Orthogonal LoRA Composition.
For task $t$, any parameter matrix (e.g.~different $\mathbf{W}_Q, \mathbf{W}_K, \mathbf{W}_V$ within different transformer blocks) in ViT is denoted by $\mathbf{W}_t$ for simplicity, and it can be formulated as a composition of base model parameters and sequentially learned LoRA matrices: 
\begin{equation}
    \begin{split}
    \mathbf{W}_{t} &=\mathbf{W}_0+\omega_1\Delta \mathbf{W}_1+\omega _2\Delta \mathbf{W}_2+\cdots +\omega _{t}\Delta\mathbf{W}_{t},
    \end{split}
    \label{eq:lorac}
\end{equation}

\noindent where $\Delta \mathbf{W}_t=\mathbf{A}_t\mathbf{B}_t\in \mathbb{R}^{K\times D}$, $\mathbf{A}_{t}\in \mathbb{R}^{K\times R}$ and $\mathbf{B}_{t}\in \mathbb{R}^{R\times D}$. $\mathbf{W}_0\in \mathbb{R}^{K\times D}$ denotes the pre-trained parameter matrix. 
We assign weights $\boldsymbol{\omega}=(\omega_1, \omega_2, \cdots, \omega_{t})$ to different LoRA matrices. 
When learning task $t$, the recently incorporated LoRA module $\Delta \mathbf{W}_t=\mathbf{A}_t\mathbf{B}_t$ and the weight coefficients $\boldsymbol{\omega}$ are updated, while the previously learned LoRA modules $\Delta \mathbf{W}_{\tau}, (\tau=1,2,\cdots,t-1)$ are kept frozen to preserve the knowledge acquired from pre-tasks.

It is evident that a proficient continual learner has the capability to acquire diverse knowledge across different tasks with minimal interference between existing and new knowledge.
 To achieve this, we strive to impose orthogonality regularization on the learned weights (\ie{} $\Delta \mathbf{W}_{\tau}, \tau=1,2,\cdots,t$) for different tasks.
The recent research, LoRA-FA~\citep{lora-fa}, suggests performing QR decomposition on the projection-down weight of $\mathbf{A}_t$. Thus, the change of model weights $\Delta \mathbf{W}_t$ will be constrained in a low-rank space as follows:
\vspace{-10pt}
\begin{align}
    \Delta \mathbf{W}_t&=\mathbf{A}_t\mathbf{B}_t,   \nonumber\\ 
    &=\mathbf{Q}_t\mathbf{R}_t\mathbf{B}_t, \nonumber\\ 
    &=\mathbf{Q}_t\mathbf{K}_t,
\end{align}

\noindent where $\mathbf{Q}_t\in \mathbb{R}^{K\times R}$ and the $R$ columns of $\mathbf{Q}_t$ are orthogonal unit vectors. $\mathbf{R}_t\in \mathbb{R}^{R\times R}$ represents right triangular matrix. We denote $\mathbf{K}_{t}=\mathbf{R}_t\mathbf{B}_t$, and consequently, can derive that:


  
  
  


\begin{equation}
\label{wbq}
    \Delta \mathbf{W}_t = \mathbf{Q}_t\mathbf{K}_t,\quad \mathbf{w}_j=\sum_{i=1}^R{k_{ij}\mathbf{q}_i},
\end{equation}
where \noindent $\mathbf{w}_j$ and $\mathbf{q}_i$ denote the $j$-th column and the $i$-th column of $\Delta \mathbf{W}_t$ and $\mathbf{Q}_t$, respectively. $k_{ij}$ represents the element located at the $(i,j)$-th position in $\mathbf{K}_t$. The formula above indicates that every column vector in $\Delta \mathbf{W}_t$ is formed as a linear combination of the column vectors in $\mathbf{Q}_t$. 


Our primary objective is to achieve orthogonality between $\Delta \mathbf{W}_{\tau}$ $(\tau=1,2,\cdots,t-1)$ and $\Delta \mathbf{W}_{t}$. 
Based on the above discussion, we can impose the constraint that the column vectors of $\mathbf{Q}_1,...,\mathbf{Q}_{t-1}$ and $\mathbf{Q}_t$ are orthonormal.  we concatenate $\mathbf{Q}_1,...,\mathbf{Q}_t$ by columns:

\begin{equation}
    \tilde{\mathbf{Q}}_t=\left[ \mathbf{Q}_1,\mathbf{Q}_2,\cdots ,\mathbf{Q}_{t} \right] 
\end{equation}
\noindent where $\tilde{\mathbf{Q}}_{t}\in \mathbb{R}^{K\times \left(t \times R\right)}$.
Thus, the following losses are formulated to enforce orthogonality on sequentially learned LoRA matrices. 
\begin{equation}
\mathcal{L}_{\mathrm{ortho}}(\tilde{\mathbf{Q}}_t) = \lVert \tilde{\mathbf{Q}}_t^\top \tilde{\mathbf{Q}}_t-\mathbf{I} \rVert _2
\label{Eq:ortho}
\end{equation}
\noindent Here, $\mathbf{I}$ represents the identity matrix. The proposed loss constrains that the inner product between each column vector of $\tilde{\mathbf{Q}}_t$ and itself is 1, while the inner product with other column vectors is 0. 
During the learning of task $t-1$, the column vectors of $\mathbf{Q}_1,...,\mathbf{Q}_{t-1}$  are constrained to be orthogonal to each other, which remain frozen in subsequent tasks. Thus, when learning task $t$, this constraint enforces that the column vectors of $\mathbf{Q}_t$  are orthonormal and mutually orthonormal with the column vectors of $\mathbf{Q}_1,...,\mathbf{Q}_{t-1}$.


    

     
    

\subsection{Important Parameter Constraints}
\label{sec:ipc}
The orthogonal regularization enforces the currently learned LoRA matrix to be orthogonal to the previously learned LoRA matrices $\Delta \mathbf{W}_{\tau}$ $(\tau=1,2,\cdots,t)$. 
From the principle of orthogonal gradient descent (OGD)~\citep{ogd}, the training loss of the previous tasks may not change notably, as updating the parameters along the direction orthogonal to the gradient would not change the loss.
Even though we use low-rank adaption, which may not guarantee strict orthogonality, the loss associated with the previous task would not change notably.

However, in this work, we find the parameters sensitive to previous training loss change observably, as shown in Fig.~\ref{fig:figure1} (b). 

Specifically, the first column of Fig.~\ref{fig:figure1} (b) illustrates the average importance of each parameter matrix for different tasks.
The importance is determined based on the sensitivity of parameters to training loss, with calculation details provided later in this section.
The darker squares in the plot indicate that the parameter matrix is more sensitive to the training loss of task $t$, i.e., the parameter matrix is more important.
The right two columns represent the variation in each parameter matrix following the model's completion of task $t+1$ and all subsequent tasks, respectively;  darker squares indicate a greater change in the parameter matrix. 
As demonstrated in the figure, the model parameters sensitive to previous training losses (marked with yellow boxes) change observably.

Besides, we further verify the orthogonality of LoRA matrices.
The results in Fig.~\ref{fig:figure1} (a) show the columns of $\mathbf{Q}_\tau$ are orthonormal, and its column vectors are mutually orthonormal to those of other matrices. We speculate this to the fact that in extremely high-dimensional parameter spaces, low-rank orthogonal solutions do not guarantee that the original parameter matrices are orthogonal to each other. 

To better regularize the final solution, we propose the Important Parameter Constraints (IPC). 
Following the previous work~\citep{platon}, which assesses the importance of model parameters according to the loss change when a parameter is zeroed out, we first define the importance of a trainable parameter $w_{t,ij}$ for task $t$

\begin{equation}
    I\left( w_{t,ij} \right) =\left| w_{t,ij}\nabla _{w_{t,ij}}\mathcal{L} \right|.
    \label{eq:ipc_base}
    \vspace{-1pt}
\end{equation}
where $\mathcal{L}$ represents the training loss of the model for task $t$, and $\nabla _{w_{t,ij}}\mathcal{L}$ indicates the gradient of the loss with respect to the training parameter $w_{t,ij}$. Thus, the importance score $I\left( w_{t,ij} \right)$ for the trainable parameter $w_{t,ij}$ is defined as the product of the parameter and its corresponding gradient.

As the importance score is calculated in the sampled mini-batch, 
the stochastic sampling and complex training dynamics result in high variability and uncertainty in the calculation.
Thus, we use sensitivity smoothing and uncertainty quantification as in~\citep{platon} to alleviate this problem.
\begin{equation}
    \begin{aligned}
        \bar{I}\left( w_{t,ij} \right) &=\beta _1\bar{I}\left( w_{t,ij} \right) +\left( 1-\beta _1 \right) I\left( w_{t,ij} \right)\\
        \bar{U}\left( w_{t,ij} \right) &=\beta _2\bar{U}\left( w_{t,ij} \right) +\left( 1-\beta _2 \right) U\left( w_{t,ij} \right)\\
    \end{aligned}
    \label{eq:ipc_shift}
\end{equation}
\noindent where 
$I\left( w_{t,ij} \right)$ is the sensitivity-based importance of parameter $w_{t,i j}$ and $\bar{I}\left( w_{t,ij} \right)$ is the smoothed sensitivity-based importance by exponential moving average.
$U\left( w_{t,ij} \right) =|I\left( w_{t,ij} \right) -\bar{I}\left( w_{t,ij} \right)|$ is the uncertainty term quantified by the local variation between $I\left( w_{t,ij} \right)$ and $\bar{I}\left( w_{t,ij} \right)$, and $\bar{U}\left( w_{t,ij} \right)$ is the smoothed result obtained by applying the exponential moving average.
$\beta _1>0,\beta _2<1$ are adjustable hyperparameters.
Then the importance of the parameter $w_{t,ij}$ is defined as the product between $\bar{I}\left( w_{t,ij} \right)$ and $\bar{U}\left( w_{t,ij} \right)$:
\begin{equation}
    S\left( w_{t,ij} \right) =\bar{I}\left( w_{t,ij} \right) \cdot \bar{U}\left( w_{t,ij} \right) 
    \label{eq:ipc}
\end{equation}
\noindent 
Then, we define the average importance of the parameter matrix for task $t$:
\begin{align}
S\left( \mathbf{W}_{t} \right)&=S\left( \mathbf{W}_{0}+\omega _{1}\Delta \mathbf{W}_{1}+\omega _{2}\Delta \mathbf{W}_{2}+\cdots +\omega _{t}\Delta \mathbf{W}_{t} \right) \\
&=\frac{1}{K\times D}\sum_{i=1}^K{\sum_{j=1}^D{S\left( w_{t,ij} \right)}}  
\label{eq:postion_ipc}
\end{align}
where $w_{t, ij}$ represents the element of the $i$-th row and $j$-th column of $\mathbf{W}_{t}$. 
The larger $S\left( \mathbf{W}_{t} \right)$ is, the more important the parameter matrix is for the current task.
The upper part of Fig.~\ref{fig:lorac-ipc} illustrates the workflow of Important Parameter Constraints. After the model finishes training on current task, we calculate the average parameter importance for each parameter matrix across different blocks in ViT. These parameter matrices are then sorted by their importance, from highest to lowest. The top-p most important matrices are selected for freezing before learning subsequent tasks, thereby maintaining the model’s performance on current task to a certain extent.

\begin{figure}[tp]
    \centering
    \includegraphics[width=0.98\linewidth]{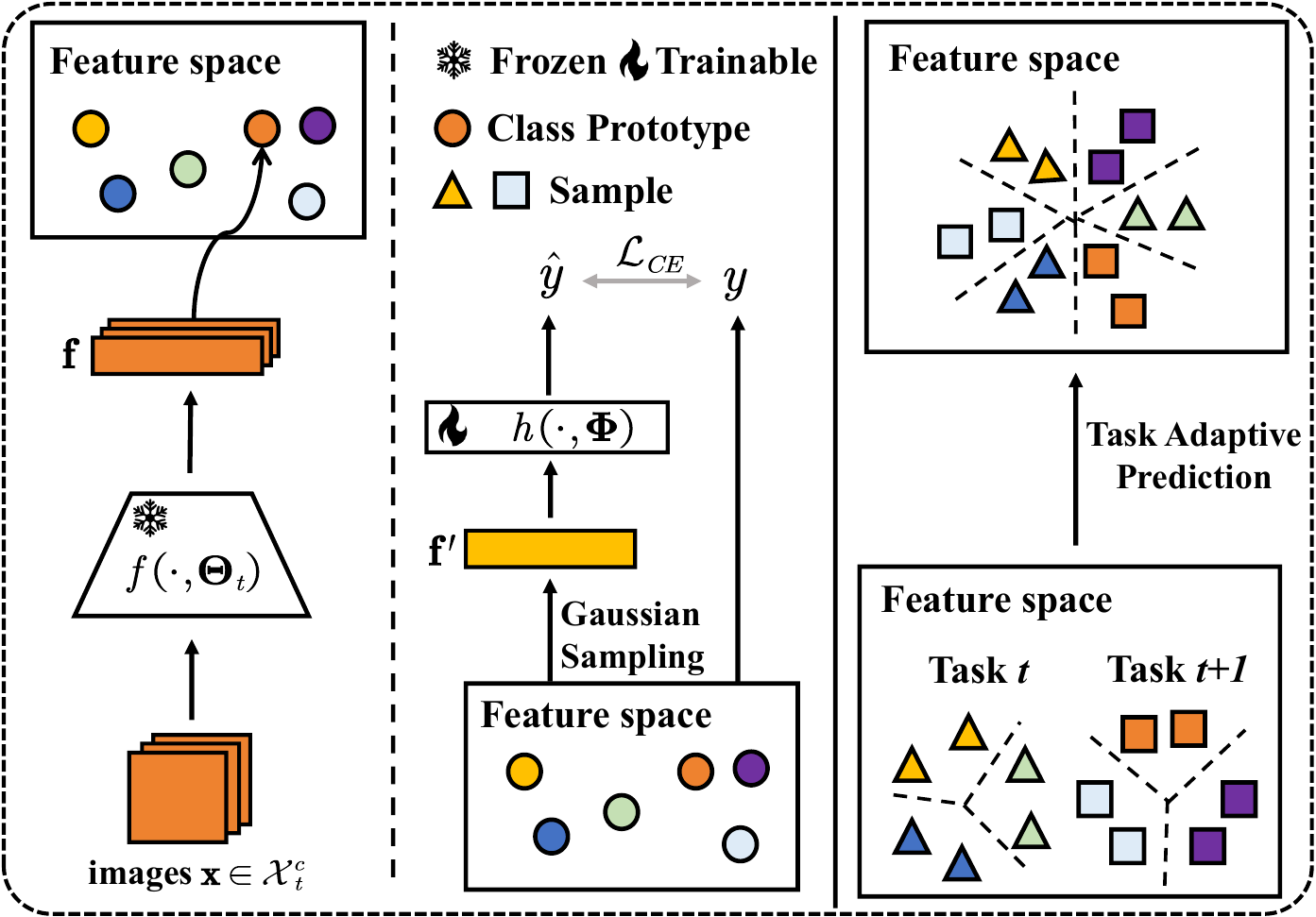}
    \caption{Parameter Adjustment for Task Adaptive Prediction. We use the feature extractor $f\left(\cdot,\mathbf{\Theta}_t\right)$, trained on task $t$, to extract the prototypes of each class in that task. Then, we perform Gaussian sampling from the class prototypes to obtain the pseudo features $\mathbf{f}'$ for adjusting the classifier $h(\cdot, \boldsymbol{\Phi})$. After completing Task Adaptive Prediction, the classifier can distinguish classes from different tasks.} 
    \label{fig:tap}
    \vspace{-10pt}
\end{figure}

\subsection{Continual Learning with Orthogonal LoRAC and IPC}
The total loss function for fine-tuning on each task $t$ is expressed as:
\begin{equation}
\label{Eq:FineTune}
\mathcal{L}\left(\bm{\Theta}_t,\bm{\Phi}_t\right)=\mathbb{E}_{\mathcal{D}_t}\left[\mathcal{L}_{\mathrm{CE}}\left(h\left(f\left(\mathbf{x},\bm{\Theta}_t \right), \bm{\Phi}_t\right),y\right)\right] + \lambda
\mathcal{L}_{\mathrm{ortho}}(\tilde{\mathbf{Q}}_t )
\end{equation}

where $\mathcal{L}_{\mathrm{CE}}(\cdot, \cdot)$ is a cross-entropy (CE) loss, and $\lambda$ is a hyperparameter used to balance the last term. 
In the learning process, we also sequentially freeze the top-p important parameter matrices
for pre-task to update $\bm{\Theta}_t$ and $\bm{\Phi}_t$.
In addition, we also update exclusively the weight coefficients $\omega_1, \omega_2, \cdots, \omega_{t-1}$ with a low learning rate to slightly relax orthogonality.
This may promote the plasticity of our method.

\subsection{Parameter Adjustment for Task Adaptive Prediction}
In the above learning process, the parameter matrix $\mathbf{\Phi}_t$ for each task classifier is learned only on the examples in that task.  
After completing task $T$, $\bm{\Phi}=\left[\bm{\Phi}_1,...,\bm{\Phi}_T\right]$ should be adjusted for the examples from all tasks.
Thus, we frozen the feature extractor and sample the pseudo features equally from a series of Gaussian distributions, each of which is centered at $\hat{\boldsymbol{\mu}}_{c}$, with the covariance estimated on each task's real data.
Here, $\hat{\boldsymbol{\mu}}_{c}=\frac{1}{|\mathcal{X}^{c}_{t}|}\sum_{\mathbf{x}\in\mathcal{X}_{t}^{c}}{f\left(\mathbf{x}, \mathbf{\Theta}_{t}\right)}$, which is the class prototype for the class $c$ belong to task $t$.
Let $\mathbf{f}'$ denote the pseudo feature and $\mathcal{D}_t'$ denote the sampled dataset, the classifier $h(\cdot, \boldsymbol{\Phi})$ could be adjust by optimizing the following objective:
\begin{equation}
\mathcal{L}'\left(\bm{\Phi}\right) = \frac{1}{T}\sum_{t=1}^T \mathbb{E}_{\mathcal{D}'_t}[\mathcal{L}_{\text{CE}}(h(\mathbf{f}', \boldsymbol{\Phi}), y)]\label{Eq:refine}
\end{equation}
Fig.~\ref{fig:tap} illustrates such parameter adjustment for task adaptive prediction.

\section{Inference}
In inference, to extract more effective representation, we first predict a test example's task ID $t^{*}$ and use $\mathbf{\Theta}_{t^*}$ for feature extraction.  Then, we predict its class through the classifier $h\left(\cdot, \bm{\Phi}\right)$, where $\bm{\Phi}$ is adjusted for task adaptive prediction.
\subsection{Task ID Inference}

Following the recent studies~\citep{RanPAC}, we use the feature extractor $f(\cdot,\mathbf{\Theta}_{1})$ to extract the class prototype $\boldsymbol{\mu}_{c}=\frac{1}{|\mathcal{X}_{t}^{c}|}\sum_{\mathbf{x}\in\mathcal{X}_{t}^{c}}{f\left(\mathbf{x}, \mathbf{\Theta}_{1}\right)}$ and its corresponding covariance $\boldsymbol{\Sigma}_{c}$, where $c\in\mathcal{Y}_{t}$ and $t\in[1,T]$. 
As analyzed in these works, PTMs adapted in the first task exhibit a decent quality of representation for samples from each task due to bridging the domain gap.
Then, based on Nearest Class Mean (NCM), we predict the task ID of the example $\mathbf{x}$ as follows:
\begin{equation}
    t^{*} = \underset{t}{\operatorname{argmin}}\{D_{M}(\mathbf{f}, \boldsymbol{\mu}_{c}) | c\in\mathcal{Y}_{t}, t\in[1,T]\}
\end{equation}
where $\mathbf{f}=f(\mathbf{x},\mathbf{\Theta}_{1})$ and the squared Mahalanobis distance is defined as $D_{M}(\mathbf{f}, \boldsymbol{\mu}_{c})=(\mathbf{f}-\boldsymbol{\mu}_{c})^{\top}\boldsymbol{\Sigma}_{c}^{-1}(\mathbf{f}-\boldsymbol{\mu}_{c})$.
Then, the representation for $\mathbf{x}$ is obtained by $f\left(\mathbf{x},\mathbf{\Theta}_{t^*}\right)$.

\subsection{Class Inference}
Finally, we predict the class of the example $\mathbf{x}$ using the adjusted classifier, $y^{}=h(f(\mathbf{x}, \mathbf{\Theta}_{t^*}), \boldsymbol{\Phi})$, where the feature representation is extracted by $f(\mathbf{x}, \mathbf{\Theta}_{t^*})$. Task-specific parameters $\mathbf{\Theta}_{t^*}$ are used for more effective representation calculation.

\section{Experimental Results}
\label{sec:Exp}

\begin{figure*}[ht!]
    \centering
    \includegraphics[width=0.98\linewidth]{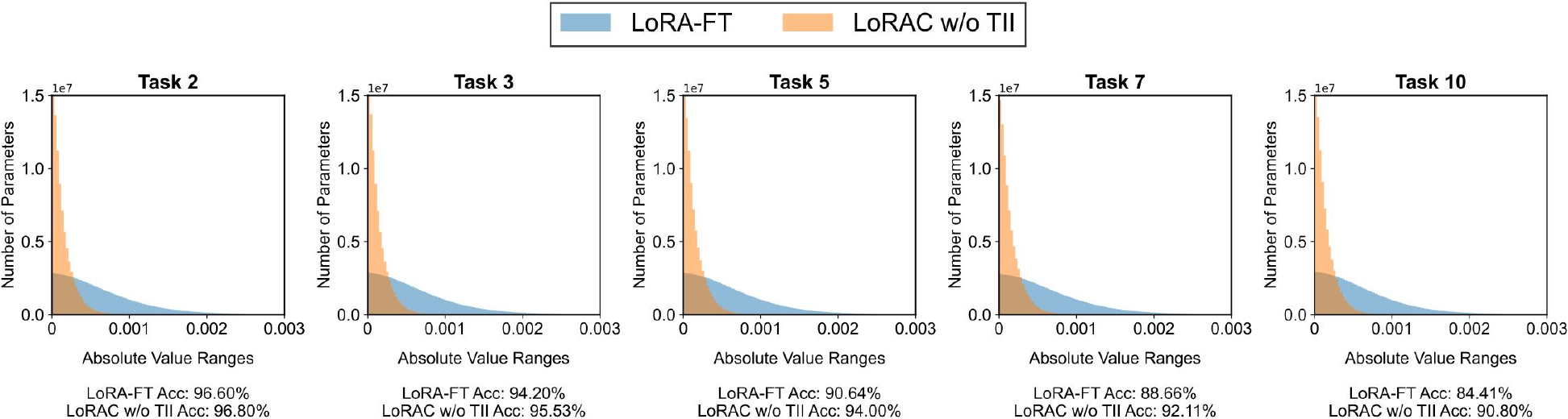}
    \caption{Delta parameter absolute values of the model on each task. Based on the Sup-21k* pre-trained model learned sequentially on Split CIFAR-100 using LoRA-FT and \FrameworkName{} w/o TII, respectively, the variations of the model's parameter with LoRA on tasks 2, 3, 5, 7, and 10 are shown, along with the average accuracy.}
    \vspace{-10pt}
    \label{fig:delta_param}
\end{figure*}

\subsection{Experimental Setup}
\subsubsection{Datasets}

\noindent \textbf{Split CIFAR-100.} To create the Split CIFAR-100 dataset, we split the CIFAR-100 dataset~\citep{cifar100} into 10 tasks. Each task contains 10 classes, each of which has 500 and 100 images of size $32\times 32$ for training and testing, respectively.

\noindent \textbf{Split ImageNet-R.} 
ImageNet-R~\citep{imagenet-r} dataset has renditions of 200 ImageNet~\citep{ridnik2021imagenet} classes resulting in 30,000 images of size $256 \times 256$. The Split ImageNet-R dataset is a modified version of the ImageNet-R dataset with 200 classes divided into 10 tasks. Each task is composed of 20 separate classes.

\noindent \textbf{5-datasets.}
The 5-datasets~\citep{5-datasets} is a combination of five datasets, namely SVHN, MNIST, CIFAR-10, Not-MNIST, and Fashion-MNIST. 
Images in the CIFAR-10 and SVHN datasets have a size of 32x32, while those in the MNIST, Fashion-MNIST, and Not-MNIST datasets have a size of 28x28.
Each of these datasets is treated as an incremental task to evaluate the impact of large inter-task differences.

\noindent \textbf{Split DomainNet.}
 DomainNet~\citep{domainnet} is a cross-domain dataset, including 345 classes and 409,832 images. 
Since images come from different domains, their original sizes vary. For ease of use, these images are typically resized to standard size, such as $224\times224$ or $256\times256$.~\citep{wang2022sprompts, smith2023coda}
 This dataset is considered more challenging due to its large number of classes and significant disparity in image counts across different classes~\citep{cpompt}. Following existing continual learning works~\citep{smith2023coda, inflora}, we split DomainNet into 5 tasks, each containing 69 classes for class-incremental learning.

\subsubsection{Evaluation Metrics}
To evaluate the performance of our model, we use the test accuracy of task  $\mathcal{T}_{\tau}$ after learning task $\mathcal{T}_{i}$ denoted by $a_{i,\tau}$ and follow the widely used incremental metrics: \textbf{Average Accuracy (Avg. Acc)} and \textbf{Average Forgetting (Forget)}.
 Main experimental results are averaged over 3 runs, and the corresponding standard deviation is reported.

\noindent \textbf{Average Accuracy (Avg. Acc)} is the average test accuracy of all tasks after the model has been trained on the task $T$ in sequence. It is defined as: $\mathbf{A}=\frac{1}{T}\sum_{\tau =1}^T{a_{T,\tau}}$.

\noindent \textbf{Average Forgetting (Forget)} is the drop in task performance averaged over previous tasks. It refers to the average decrease in performance for each task from its maximum accuracy to the accuracy achieved at the completion of training. It is defined as follows: $\mathbf{F}=\frac{1}{T-1}\sum_{\tau =1}^{T-1}{\max _{i\in \left\{ 1,...,T-1 \right\}}\left( a_{i,\tau}-a_{T,\tau} \right)}$.

\subsubsection{Implementation Details}

Following similar implementations as previous work~\citep{wang2022learning, wang2022dualprompt, wang2023hide,cpp,zhang2023slca}, we mainly
utilize three PTMs (based on ViT-B/16~\citep{vit}): one with supervised pre-trained on ImageNet-21K (denoted as Sup-21K), another with data augmentation (denoted as Sup-21K*), and one with self-supervised pre-trained on ImageNet-1K (denoted as MoCo-1K). 

We adopt the Sup-21K backbone and train using the Adam optimizer with a batch size of $128$.
For Split CIFAR-100, Split ImageNet-R, and 5-datasets, the learning rate is set to $0.02$, $0.01$, and $0.006$, respectively.
The sizes of the input images are adjusted to 224 × 224.
When performing the important parameter constraints, we set the hyperparameters $\beta_1$ and $\beta_2$ as their default value 0.85. 
The parameter matrices with average importance in the top 5\% or 10\% are empirically selected as important for the current task and are frozen before learning the subsequent tasks to minimize forgetting.
Please refer to Appendix A for more experimental details.

\subsection{Orthogonal LoRA Composition Analysis}
We analyze the validity of orthogonal LoRA compositions by examining the delta parameters' absolute values.
The delta parameter's absolute value is defined as $\left|\Delta w\right|=\left| w_{t,ij}-w_{t-1,ij} \right|$, where $w_{t,ij}\in \mathbf{W}_t$ is an element of the \emph{i}-th and \emph{j}-th column of $\mathbf{W}_t$ and $\mathbf{W}_t$ can be denoted as the parameter matrix for each transformer blocks in the ViT model after the model has learned task $t$.
As illustrated in \Cref{fig:delta_param}, 
we conduct a statistical analysis of the elements of all parameter matrices in all transformer blocks of the ViT model for each task.
LoRA-FT represents the fine-tuning of a pre-trained model by initializing a new LoRA for each task without any regularization, using it to acquire task-specific knowledge. 
 
The results show two notable trends. 
First, on each task, the delta parameter absolute values of LoRA-FT exhibit a distribution range of $0$ to $0.002$. In contrast, most delta parameter absolute values of \FrameworkNameWo{} are concentrated within $0$ to $0.0005$. This indicates that \FrameworkNameWo{} induces fewer changes to the model parameters for each task.
 Second, as the number of tasks increases, the difference in average accuracy between LoRA-FT and \FrameworkNameWo{} gradually grows. For instance, on Task 10, \FrameworkNameWo{} achieves a $6.39\%$ higher average accuracy than LoRA-FT.
We attribute these findings to the fact that \FrameworkNameWo{} learns on each task by composition, which combines the knowledge learned on previous tasks, and under the orthogonal loss constraint, the changes to the parameters are much more slight. 
This reduces interference with parameters learned on previous tasks, ensures a high degree of model stability, and maintains accuracy for prior and current tasks.

\begin{figure*}[ht!]
    \centering
    \includegraphics[width=0.98\textwidth]{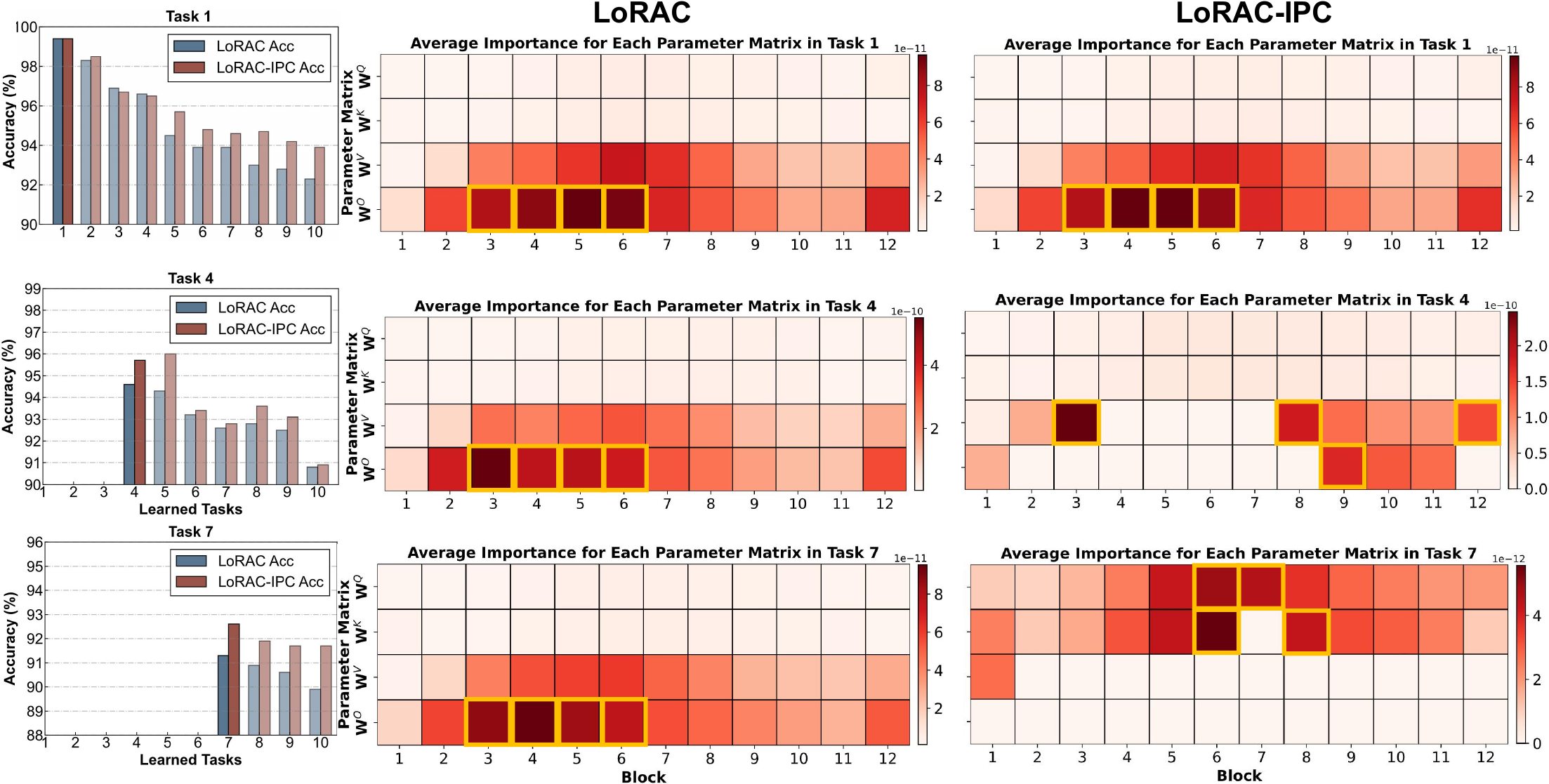}
    \caption{
    The bar graphs on the left depict the variation in accuracy of LoRAC and LoRAC-IPC across various tasks on Split CIFAR-100. 
    The right half shows important parameters for the current tasks of LoRAC and LoRAC-IPC.
    Here we select the parameter matrices in the top 10\% of importance for the current task.
    Important parameters for each task are highlighted by yellow boxes. 
    }
    \label{fig:ipc_anlysis}
\end{figure*}

\subsection{Important Parameter Constraints Anlysis}

Recall in Sec. \ref{sec:ipc}, we discussed even if we impose orthogonal regularization on LoRA matrices, the positions of important parameters also overlap largely. 
To validate the effect of important parameter constraints, we show the important parameter position of LoRAC before and after using important parameter constraints.

The bar graphs on the left side of \Cref{fig:ipc_anlysis} depict the change in the model's accuracy across different tasks on Split CIFAR-100. 
The right side illustrates the average parameter importance for each parameter matrix across different blocks of the model for various tasks before and after using important parameter constraints(IPC).
From \Cref{fig:ipc_anlysis}, it can be observed that 
(1) With IPC, the positions of important parameters do not overlap largely. 
(2) The degradation of the model's performance on the current task during subsequent continual learning is effectively mitigated after imposing IPC.
This suggests that IPC is effective in mitigating the model's forgetting of knowledge related to the current task by freezing the most important parameter matrices. 
This further demonstrates the necessity and effectiveness of IPC.

\begin{table*}[ht!]
\begin{center}
\caption{Results for rehearsal-free continual learning on Split CIFAR-100 and Split ImageNet-R. 
Sup-21K: supervised pre-training on ImageNet-21K. 
Sup-21K*: supervised pre-training on ImageNet-21K with data augmentation. 
MoCo-1K: self-supervised pre-training on ImageNet-1K with MoCo v3.
$\dag$ Used checkpoints fine-tuned from Sup-21K* on ImageNet-1K.
$^{\ddag}$ Reproduced using their original codebases after revision.
For Batch-Wise testing, multiple test samples with the same task ID constitute a batch. The best results are highlighted in bold, while the second-best results are underlined.
}
\resizebox{0.75\linewidth}{!}{%
\begin{tabular}{llclclc}
	 \toprule 
        \multirow{2}{*}{\textbf{PTM}} & \multirow{2}{*}{\textbf{Method}} & \multirow{2}{*}{\textbf{Batch-Wise}} & \multicolumn{2}{c}{\textbf{Split CIFAR-100}}  & \multicolumn{2}{c}{\textbf{Split ImageNet-R}} \\
        && & Avg. Acc ($\uparrow$) & Forget ($\downarrow$)  & Avg. Acc ($\uparrow$) & Forget ($\downarrow$) \\
        \midrule
       \multirow{9}*{Sup-21K} 
       &\emph{Joint-Training} & & 93.15\scriptsize{$\pm$0.09} & - & 83.87\scriptsize{$\pm$0.30} & - \\
       \cline{2-7}
       &\emph{Seq-FT} & & 17.72\scriptsize{$\pm$0.34} & 59.09\scriptsize{$\pm$0.25} & 28.87\scriptsize{$\pm$1.36} & 63.80\scriptsize{$\pm$1.50} \\
      &LAE~\citep{lae}& &85.59\scriptsize{$\pm$0.46} &-  &72.66\scriptsize{$\pm$0.63} &-\\       
       &L2P~\citep{wang2022learning}&\checkmark &86.31\scriptsize{$\pm$0.59} &5.83\scriptsize{$\pm$0.61} &61.57\scriptsize{$\pm$0.66} &9.73\scriptsize{$\pm$0.47} \\ 
       &DualPrompt~\citep{wang2022dualprompt} &\checkmark &86.51\scriptsize{$\pm$0.33} &5.16\scriptsize{$\pm$0.09} 
       &68.13\scriptsize{$\pm$0.49} 
       &4.68\scriptsize{$\pm$0.20} \\ 
       &HiDe-Prompt~\citep{wang2023hide}$^{\ddag}$ &&85.48\scriptsize{$\pm$0.14} 
       &5.78\scriptsize{$\pm$0.19} 
       &66.06\scriptsize{$\pm$0.05} &6.56\scriptsize{$\pm$0.38}\\ 
       &\CC{ours}{\FrameworkName}  &\CC{ours}&\CC{ours}89.82\scriptsize{$\pm$0.09} &\CC{ours}3.46\scriptsize{$\pm$0.17} &\CC{ours}73.51\scriptsize{$\pm$0.46} &\CC{ours}2.17\scriptsize{$\pm$0.40}\\
       &\CC{ours}{\FrameworkNameIPC}  &\CC{ours}&\CC{ours}\underline{90.21\scriptsize{$\pm$0.10}} &\CC{ours}\underline{2.79\scriptsize{$\pm$0.15}} &\CC{ours}\underline{74.94\scriptsize{$\pm$0.03}} &\CC{ours}\underline{1.74\scriptsize{$\pm$0.13}}\\
       &\CC{ours}{\FrameworkNameIPC} &\CC{ours}\checkmark&\CC{ours}\bf{92.86}\scriptsize{$\pm$0.11} 
       &\CC{ours}\bf{1.92}\scriptsize{$\pm$0.09}
       &\CC{ours}\bf{81.21}\scriptsize{$\pm$0.58}
       &\CC{ours}\bf{1.42}\scriptsize{$\pm$0.19}\\
       \midrule
       \multirow{16}*{Sup-21K*}
       &\emph{Joint-Training} & & 93.22\scriptsize{$\pm$0.16} & - & 81.14\scriptsize{$\pm$0.34} & - \\
       \cline{2-7}
       &\emph{Seq-FT} & & 11.60\scriptsize{$\pm$0.13} & 90.65\scriptsize{$\pm$0.03} & 14.11\scriptsize{$\pm$0.06} & 72.38\scriptsize{$\pm$0.21} \\
       &NMC~\citep{ncm}& &83.70 &- &55.56 &-\\
       &ADAM~\citep{Adam}& &87.49 &- &67.95 &-\\
       &CODA-Prompt~\citep{smith2023coda}$^\dag$ & &86.25\scriptsize{$\pm$0.74} &- &75.45\scriptsize{$\pm$0.56} &- \\
       &InfLoRA~\citep{inflora} & &87.06\scriptsize{$\pm$ 0.25} &- &75.65\scriptsize{$\pm$ 0.14} & - \\
       &CPP~\citep{cpp}$^\dag$ &&91.12\scriptsize{$\pm$0.12} &\underline{3.33\scriptsize{$\pm$0.18}}
       &74.88\scriptsize{$\pm$0.07} 
       &3.65\scriptsize{$\pm$0.03}\\
       &SLCA~\citep{zhang2023slca} & &91.53\scriptsize{$\pm$0.28} &- &77.00\scriptsize{$\pm$0.33} &-\\
       &OVOR-Deep~\citep{OVOR} & &85.99\scriptsize{$\pm$0.89} & 6.42\scriptsize{$\pm$2.03} & 76.11\scriptsize{$\pm$0.21} &7.16\scriptsize{$\pm$0.34} \\
       &DualP-PGP~\citep{dualp} & &86.92\scriptsize{$\pm$0.05} & 5.35\scriptsize{$\pm$0.19} & 69.34\scriptsize{$\pm$0.05} & 4.53\scriptsize{$\pm$0.04} \\
       &CPrompt~\citep{cpompt} & &87.82\scriptsize{$\pm$0.21} & 5.06\scriptsize{$\pm$0.50} & 77.14\scriptsize{$\pm$0.11} & 5.97\scriptsize{$\pm$0.68} \\
       &ConvPrompt~\citep{convprompt} & &88.87\scriptsize{$\pm$0.33} & 4.75\scriptsize{$\pm$0.15} & 77.86\scriptsize{$\pm$0.25} & 4.33\scriptsize{$\pm$0.24} \\
       &EASE~\citep{ease} & &87.76 & -& 76.17 & - \\
       &RanPAC~\citep{RanPAC} &&\bf{92.20} &-  &78.10 &-\\
       &\CC{ours}\FrameworkName{} &\CC{ours} &\CC{ours}91.99\scriptsize{$\pm$0.09 }&\CC{ours}2.67\scriptsize{$\pm$0.18} &\CC{ours}\underline{78.60\scriptsize{$\pm$0.37}}
       &\CC{ours}\underline{2.16\scriptsize{$\pm$0.77}} \\
       &\CC{ours}\FrameworkNameIPC &\CC{ours} &\CC{ours}\underline{92.08\scriptsize{$\pm$0.06}}&\CC{ours}\bf{2.67}\scriptsize{$\pm$0.16} &\CC{ours}\bf{79.34}\scriptsize{$\pm$0.26}
       &\CC{ours}\bf{1.78}\scriptsize{$\pm$0.44} \\
       \midrule
       \multirow{7}*{MoCo-1K}
       &\emph{Joint-Training} & & 89.11\scriptsize{$\pm$0.06} & - & 72.80\scriptsize{$\pm$0.23} & - \\
       \cline{2-7}
       &\emph{Seq-FT} & & 16.21\scriptsize{$\pm$0.25} & 89.58\scriptsize{$\pm$0.31} & 9.10\scriptsize{$\pm$0.11} & 69.67\scriptsize{$\pm$0.20} \\
       &EWC~\citep{ewc} &&81.62\scriptsize{$\pm$0.34} &- &64.50\scriptsize{$\pm$0.36} &- \\
       &LwF~\citep{li2017learning} &&77.94\scriptsize{$\pm$1.00} &- &60.74\scriptsize{$\pm$0.30} &- \\
       &SLCA~\citep{zhang2023slca} &&85.27\scriptsize{$\pm$0.08} &- &68.07\scriptsize{$\pm $0.21} &- \\
       &\CC{ours}\FrameworkName{}  &\CC{ours}&\CC{ours}\underline{85.66\scriptsize{$\pm$0.20}} 
       &\CC{ours}\underline{4.70\scriptsize{$\pm$0.10}} &\CC{ours}\underline{69.83\scriptsize{$\pm$ 0.23}} &\CC{ours}\underline{2.79\scriptsize{$\pm$0.72}}\\
       &\CC{ours}\FrameworkNameIPC  &\CC{ours}&\CC{ours}\bf{86.11}\scriptsize{$\pm$0.20} 
       &\CC{ours}\bf{3.89}\scriptsize{$\pm$0.28} &\CC{ours}\bf{70.46}\scriptsize{$\pm$ 0.41} &\CC{ours}\bf{1.84}\scriptsize{$\pm$0.27}\\
       \bottomrule
\end{tabular}
\label{tab:result_rehearsal_free}
}
\end{center}
\vspace{-16pt}
\end{table*}

\subsection{Comparison Results}

In this section, we compare \FrameworkName{} and \FrameworkNameIPC{} to the state-of-the-art rehearsal-free methods on Split CIFAR-100, Split ImageNet-R, 5-datasets, and Split DomainNet.

\Cref{tab:result_rehearsal_free} presents the performance of \FrameworkName{} and \FrameworkNameIPC{} using different PTMs: Sup-21K, Sup-21K*, and MoCo-1K. 
We also report the upper bound performance (\emph{Joint-Training}) and the lower bound performance (\emph{Seq-FT}) for these PTMs across different datasets in the table. Here, \emph{Joint-Training} denotes the method that learns all tasks jointly, while \emph{Seq-FT} denotes the method that learns all tasks sequentially without any mechanism to mitigate the model's forgetting.

\begin{figure}[ht!]

\vspace{-7pt}
    \centering
    \subfloat{
    \includegraphics[width=0.24\textwidth, page=1]{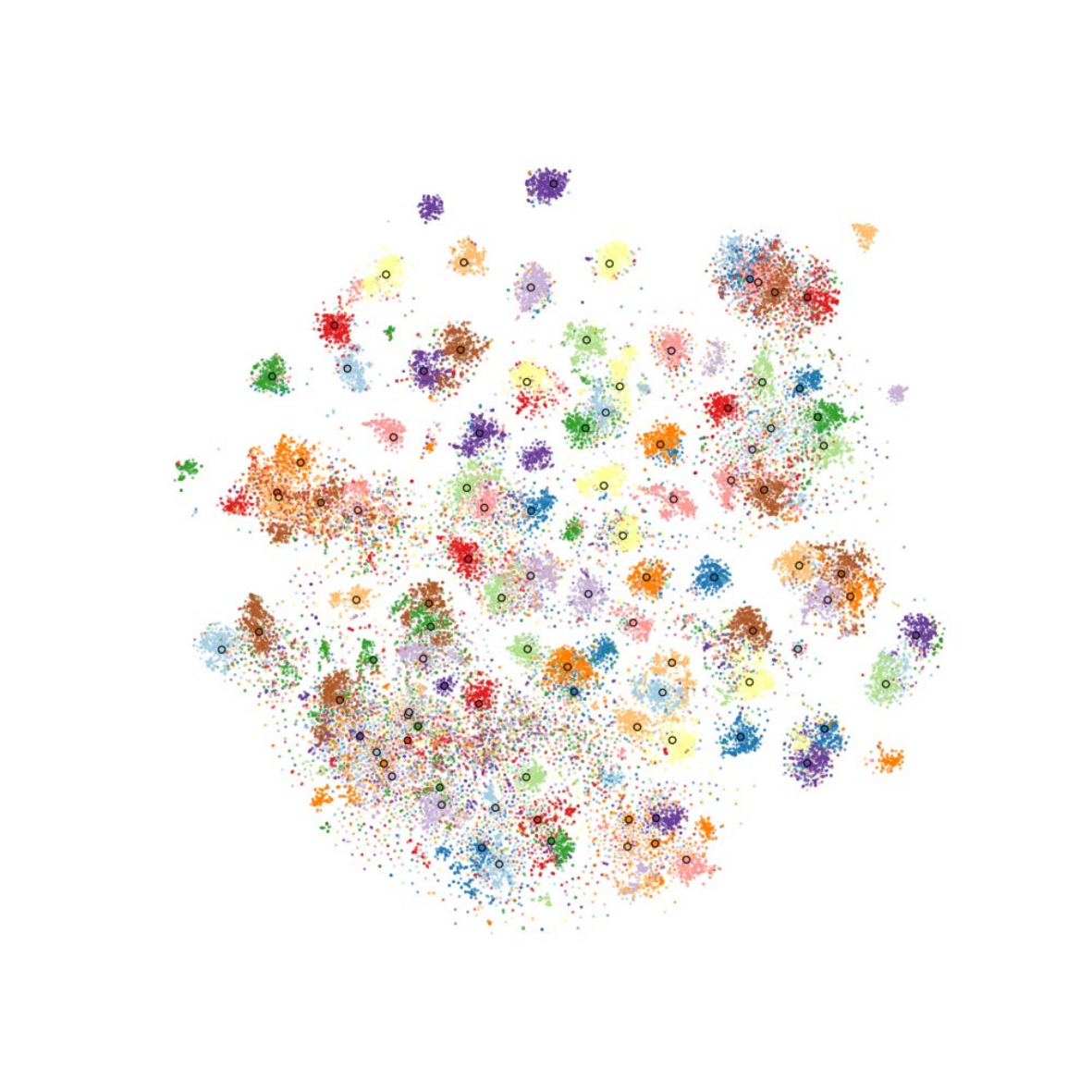}}
    \subfloat{
    \includegraphics[width=0.24\textwidth, page=2]{Figures/tsne-v2.pdf}}
    
    \caption{t-SNE visualization of representation with LoRA-FT (left) and with \FrameworkName{} (right). Each color corresponds to a different class.
    The results are obtained from the Sup-21K pre-trained model trained sequentially on Split CIFAR-100.
    }
    \label{fig:tsne_analyze}
    \vspace{-15pt}
\end{figure}

We can draw two conclusions by comparing the accuracy and forgetting across all methods.
 First, \FrameworkName{} outperforms other PTM's continual learning methods such as L2P, DualPrompt, Hide-Prompt (using Prompt), or LAE (using Adapter) on Sup-21K. 
Second, {\FrameworkNameIPC{}} achieves superior performance on both Split CIFAR-100 and Split ImageNet-R with each of the three PTMs. 
For example, when using Sup-21K on Split CIFAR-100, {\FrameworkName{}} outperforms LAE by {$4.23\%$} in average accuracy. 
This is attributed to the fact that the knowledge acquired by {\FrameworkName{}} on a new task does not interfere with the knowledge learned on previous tasks. This allows the model to more clearly distinguish the representations of samples from different classes across all tasks, as illustrated in \Cref{fig:tsne_analyze}.

\begin{table}[ht!]
\caption{Results for rehearsal-free continual learning on 5-datasets. All our experiments are conducted on Sup-21K.$\dag$ Used checkpoints fine-tuned from Sup-21K* on ImageNet-1k.$^{\ddag}$ Reproduced using their original codebases after revision. BW represents Batch-Wise testing.}
\label{tab: five-datasets}
\centering
\resizebox{0.4\textwidth}{!}{
\begin{tabular}{lcccc}
	 \toprule 
         \multirow{2}{*}{\textbf{Method}} & \multirow{2}{*}{\textbf{BW}}  & \multicolumn{2}{c}{\textbf{5-datasets}} \\
        && Avg. Acc ($\uparrow$) & Forget ($\downarrow$) \\
        \midrule
    \emph{Joint-Training}& & 97.81 & - \\
    \midrule
    \emph{Seq-FT}& & 39.49 & 42.62 \\
       EWC~\citep{ewc} & & 50.93 & 34.94 \\
       LwF~\citep{li2017learning} & & 47.91 & 38.01 \\
       L2P~\citep{wang2022learning} & \checkmark &81.14 &4.64  \\ 
       DualPrompt~\citep{wang2022dualprompt} & \checkmark & 88.08 &2.21 \\
       HiDe-Prompt~\citep{wang2023hide}$^{\ddag}$ &\checkmark&94.74 &0.21 \\ 
       CPP~\citep{cpp}$\dag$ &\checkmark&92.92  &0.19 \\
       \CC{ours}{\FrameworkName{}}  &\CC{ours}&\CC{ours}94.35 &\CC{ours}0.08\\
       \CC{ours}{\FrameworkNameIPC}  &\CC{ours}&\CC{ours}95.58 &\CC{ours}0.03\\
       \CC{ours}{\FrameworkNameIPC}  &\CC{ours}\checkmark &\CC{ours}\bf{95.77}&\CC{ours}\bf{0.01}\\
       \bottomrule
	\end{tabular}
	}
\vspace{-5pt}
\end{table}

Furthermore, {\FrameworkNameIPC{}} outperforms DualPrompt by {$6.35\%$} in average accuracy. 
Moreover, when using the more generic Sup-21K*, the gap between {\FrameworkNameIPC{}} and RanPAC on Split CIFAR-100 is only $0.1\%$. This is because RanPAC utilizes much more memory to maintain historical information, whereas we use less memory but achieve comparable performance and approach the upper bound ($93.22\%$).
On the more challenging Split ImageNet-R, \FrameworkNameIPC{} achieves $1.24\%$ higher accuracy than RanPAC.
In addition to supervised PTMs, we compare \FrameworkName{} and \FrameworkNameIPC{} with other methods on the self-supervised PTM MoCo-1K. On Split ImageNet-R, \FrameworkNameIPC{} surpasses the average accuracy of SLCA (full fine-tuning) by $2.39\%$.

\begin{figure}[ht!]
  \centering
  \subfloat{
    \centering
    \includegraphics[width=0.48\textwidth]{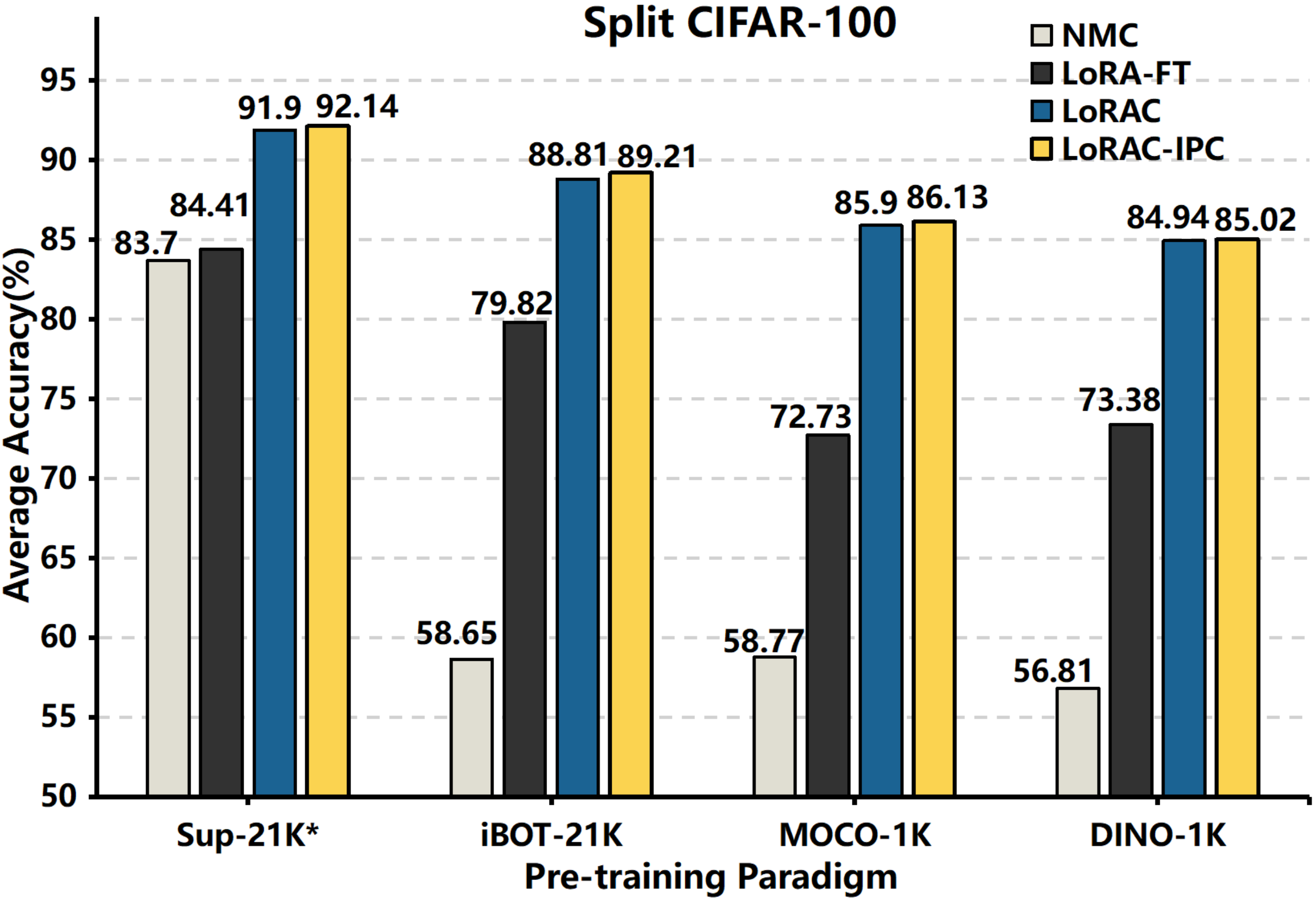}
    \label{fig:sub1}}
    \vspace{-5pt}
  \subfloat{
    \centering
    \includegraphics[width=0.48\textwidth]{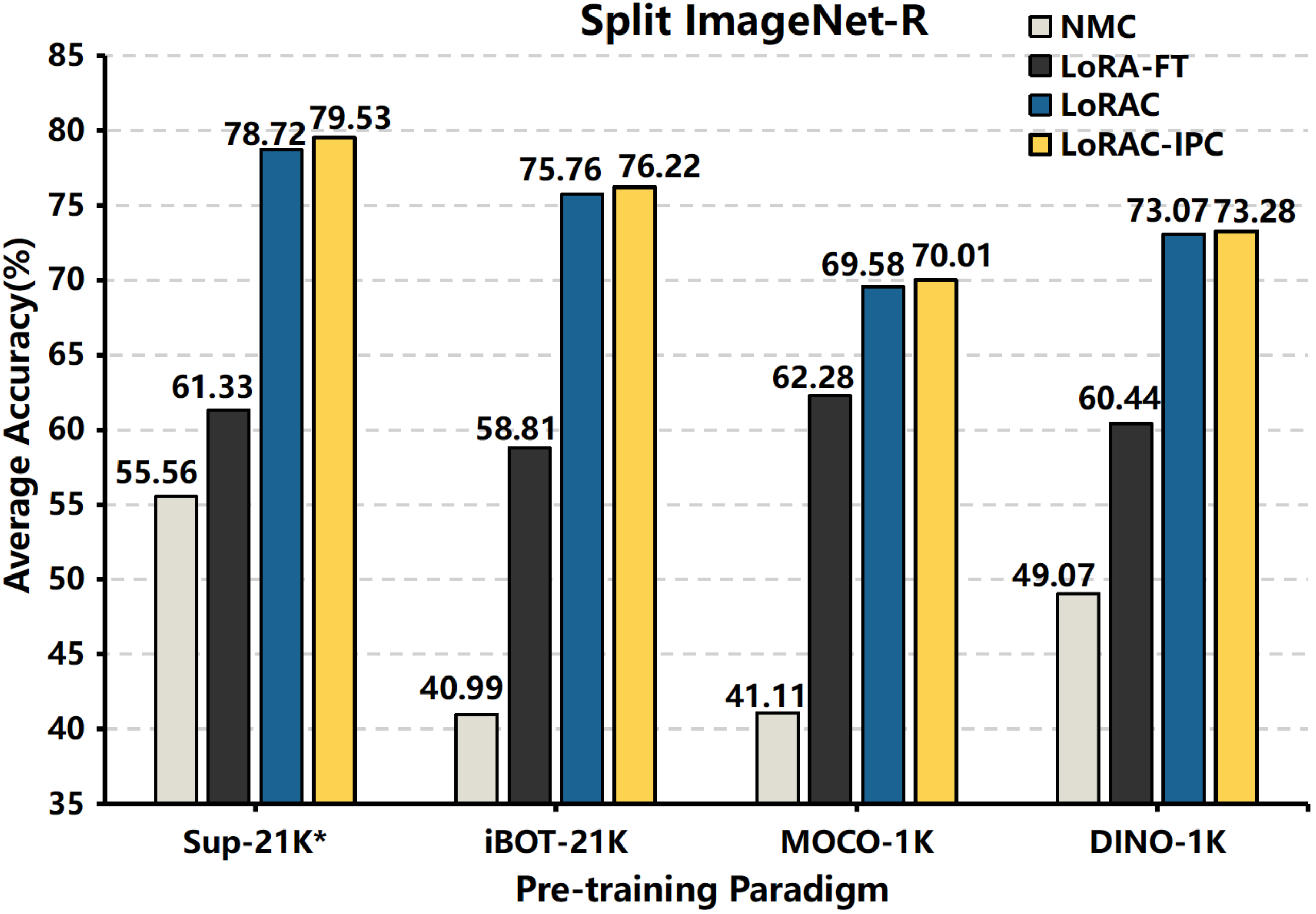}
    \label{fig:sub2}}
  \caption{
  Comparison of Nearest Mean Classifier (NMC), LoRA-FT, \FrameworkName{} and \FrameworkNameIPC{} under different pre-training paradigms.
  }
  \label{fig:ptm}
\vspace{-10pt}
\end{figure}

In \Cref{fig:ptm}, we have also conducted experiments on other self-supervised PTMs, including iBOT-21K, DINO-1K, and MoCo-1K. NMC~\citep{ncm} represents the approach of extracting image features using a frozen pre-trained ViT model and utilizing a Nearest Mean Classifier (NMC) in the feature space to make predictions on test samples. 
The figure illustrates that, with the use of self-supervised PTMs, LoRA-FT significantly outperforms NMC. \FrameworkName{} and \FrameworkNameIPC{} surpass LoRA-FT to an even greater extent.

\begin{table}[ht!]
    \centering
    \caption{Results for rehearsal-free continual learning on Split DomainNet. All our experiments are conducted on Sup-21K*.}
    \resizebox{0.4\textwidth}{!}{
    \begin{tabular}{lcc}
    \toprule
    \multirow{2}{*}{\textbf{Method}} & \multicolumn{2}{c}{\textbf{Split DomainNet}} \\
    & Avg. Acc ($\uparrow$) & Forget ($\downarrow$) \\
    \midrule
    \emph{Joint-Training} & 77.72\scriptsize{$\pm$0.04} & - \\
    \midrule
    \emph{Seq-FT} & 16.67\scriptsize{$\pm$0.01} & 83.03\scriptsize{$\pm$0.03} \\
    L2P~\citep{wang2022learning} & 70.16\scriptsize{$\pm$0.05}&- \\
    DualPrompt~\citep{wang2022dualprompt} & 72.14\scriptsize{$\pm$0.05} & - \\
    CODA-Prompt~\citep{smith2023coda} & 73.23\scriptsize{$\pm$0.13} & - \\
    C-LoRA~\citep{clora} & 69.34\scriptsize{$\pm$0.13} & - \\
    LAE~\citep{lae} & 66.85\scriptsize{$\pm$0.40} & - \\
    InfLoRA~\citep{inflora} & 74.53\scriptsize{$\pm$0.23} & - \\
    \CC{ours}{\FrameworkName{}}  &\CC{ours}75.60\scriptsize{$\pm$0.26} &\CC{ours}5.28\scriptsize{$\pm$0.11}\\
    \CC{ours}{\FrameworkNameIPC} &\CC{ours}\textbf{75.85\scriptsize{$\pm$0.15}} &\CC{ours}\textbf{5.26\scriptsize{$\pm$0.04}}\\
    \bottomrule
    \end{tabular}
    }
    \label{tab:domainnet}
    \vspace{-15pt}
\end{table}

We further verified the performance of \FrameworkName{} and \FrameworkNameIPC{} on 5-datasets with more significant differences between tasks, as shown in \Cref{tab: five-datasets}. Under Batch-Wise testing, \FrameworkNameIPC{} achieves $1.03\%$ higher accuracy than HiDe-Prompt, with forgetting $0.01\%$. This means the model has almost no forgetting under high task inference accuracy.

The experimental results for Split DomainNet are presented in \Cref{tab:domainnet}. \FrameworkNameIPC{} surpasses the state-of-the-art method in accuracy by 1.32\%, demonstrating that our proposed method remains effective on a more challenging dataset.

\subsection{Ablation Study}
This section shows our elaborate ablation study results to demonstrate the effectiveness of orthogonal LORA composition and important parameter constraints. All the experiments are conducted using Sup-21K*.

\begin{figure}[ht!]
    \centering
    \subfloat{
    \includegraphics[width=0.48\textwidth]{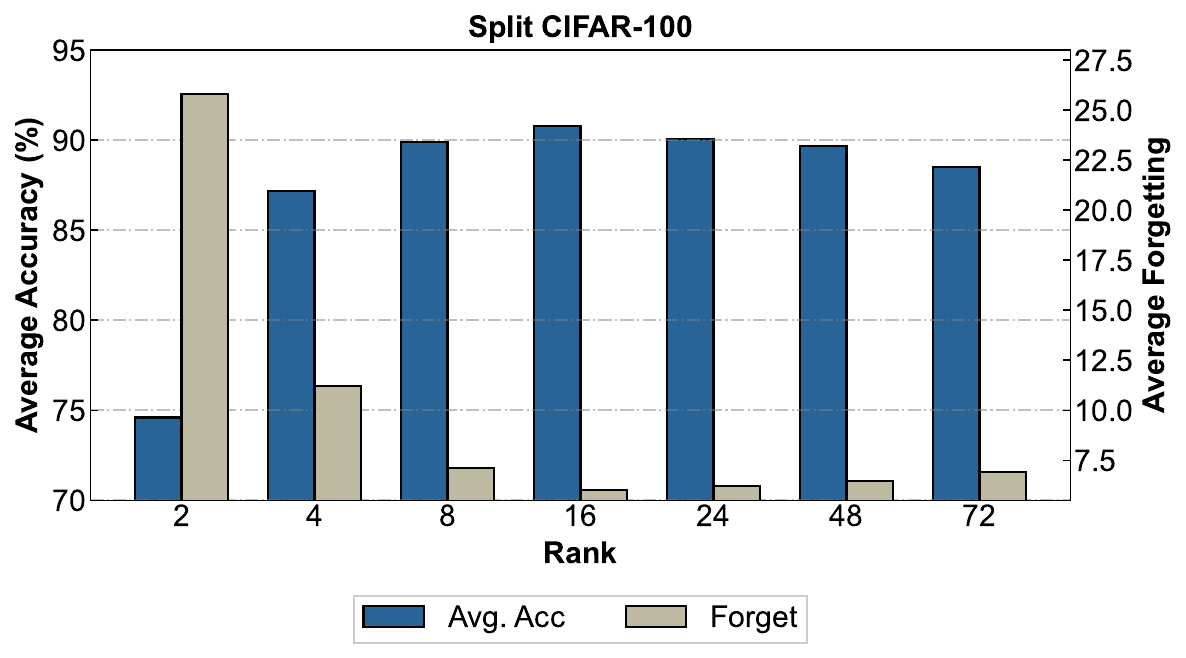}}
    \caption{Results of \FrameworkNameWo{} on Split CIFAR-100 with different rank $R$.}
    \label{fig:dis_rank}
    \vspace{-15pt}
\end{figure}

\noindent \textbf{Effect of rank R.}
Recall in Sec. \ref{sec:ortho loss}, we design the orthogonal loss constraint so that $\Delta \mathbf{W}_{\tau} (\tau=1,2,...,t-1)$ and $\Delta \mathbf{W}_t$ are orthogonal, with the goal of finding a direction orthogonal to $\Delta \mathbf{W}_{\tau} (\tau=1,2,...,t-1)$ that minimizes the impact of learning a new task on the loss of previous tasks. 
As LoRA is a low-rank approximation to full fine-tuning, the selection of the rank $R$ of the low-rank matrix is particularly crucial in this context. 
\Cref{fig:dis_rank} shows the results of \FrameworkName{} without task id inference on Split CIFAR-100 with different rank $R$. It can be seen that the performance of the model is not optimal at either lower or higher $R$, which means that the orthogonality constraints do not work best at this point; we analyze the reasons for this as follows:

Firstly, when the rank $R$ of the low-rank matrix is low, the rank of $\Delta \mathbf{W}_t$ is significantly smaller than the dimension of the pre-trained model's parameters, resulting in $\Delta \mathbf{W}_t$ poorly approximating the incremental updates under full fine-tuning. Thus, even if $\Delta \mathbf{W}_t$ is orthogonal to $\Delta \mathbf{W}_{\tau}(\tau=1,2,...,t-1)$, there is still a potential impact on the loss of the previous task, which can be mitigated by increasing the rank $R$.
Secondly, the orthogonal regularization first performs a QR decomposition of the low-rank matrix  $\mathbf{A}_{\tau}$ in $\Delta \mathbf{W}_{\tau}=\mathbf{A}_{\tau}\mathbf{B}_{\tau}$, and then constrains the orthogonality of $\mathbf{Q}_{\tau}(\tau=1,2,...,t)$, which in fact ensures that the orthogonal bases of $\Delta \mathbf{W}_{\tau}(\tau=1,2,...,t)$ are orthogonal to each other.  
Thus, when $R$ is large, the model may occupy the orthogonal bases composing the optimal solution for the subsequent task while learning the current task. 
This leads to the model achieving only sub-optimal solution in the subsequent task.
This ultimately manifests as a reduction in the model's plasticity for the subsequent task.

\begin{figure}[ht!]

        \centering
        
        \subfloat{
        \includegraphics[width=0.24\textwidth]{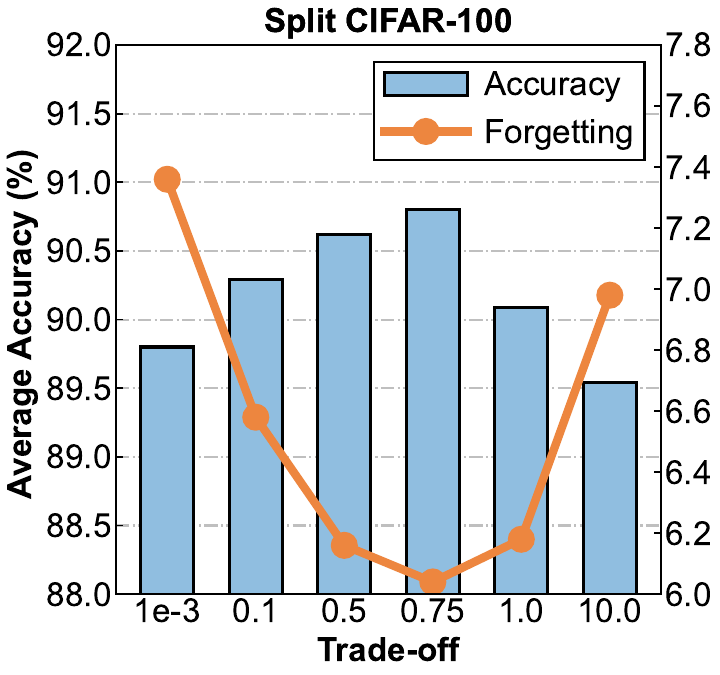}
        \includegraphics[width=0.24\textwidth]{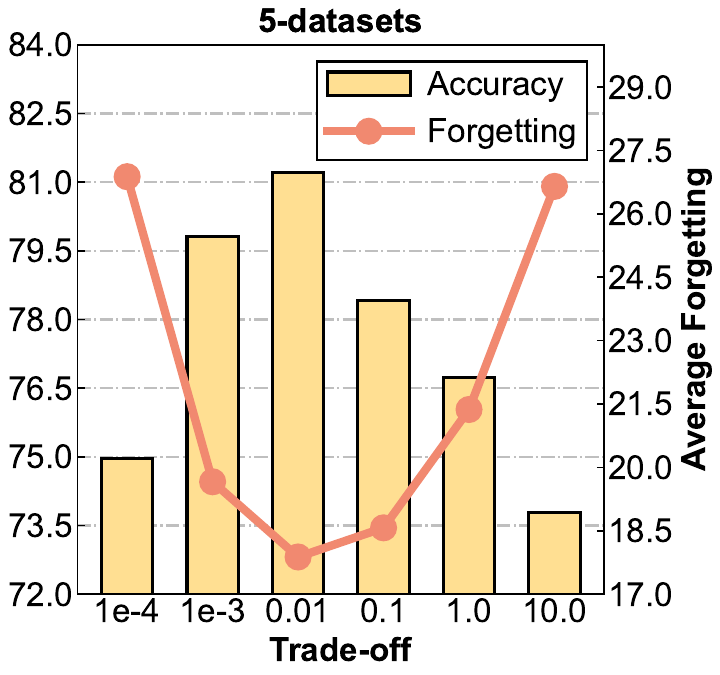}}
        \centering
    \caption{The average accuracy and forgetting of \FrameworkNameWo{} across different trade-off ($\lambda$) values for Split CIFAR-100 and 5-datasets.}
    \label{fig:trade-off}
    
\vspace{-10pt}
\end{figure}

\noindent \textbf{Effect of orthogonal loss. } \Cref{fig:trade-off} illustrates the performance of our method as the orthogonal loss trade-off value varies across both Split CIFAR-100 and 5-datasets. 
Task ID inference is not utilized in this experiment.
We plot average accuracy histograms and forgetting curves for both datasets. 
For Split CIFAR-100, the trade-off values are set to 0.001, 0.1, 0.5, 0.75, 1.0, and 10.0. While for the 5-datasets, the trade-off values are set to 0.0001, 0.001, 0.01, 0.1, 1.0, and 10. 
The figure illustrates a relatively consistent change in the performance across both datasets. 
Firstly, the forgetting curves exhibit a substantial decrease followed by a gradual increase with variations in the trade-off value. 
This observation underscores the efficacy of our proposed orthogonality regularization in mitigating catastrophic forgetting. 
Secondly, the average accuracy demonstrates an initial increase followed by a decrease. 
Our analysis suggests that the introduction of orthogonal loss initially serves to constrain the model's acquisition of knowledge for the current task, preventing interference with knowledge gained from previous tasks. 
This constraint enables the model to learn the current task while minimizing the impact of the loss of the previous task, thereby maintaining good performance on the previous task. 
But, as the trade-off values increase, the orthogonal loss reduces the optimizing impact of cross-entropy loss on the model, leading to a subsequent decline in classification accuracy.
\begin{table}[t]
\centering
\vspace{-4pt}
\caption{Comparison of different methods for selecting important parameters on Split CIFAR-100 and Split ImageNet-R.}
\label{table:ipc}
\resizebox{0.45\textwidth}{!}{
\begin{tabular}{lcccc}
    \toprule
        \multirow{2}{*}{\bf Method} &\multicolumn{2}{c}{\bf Split CIFAR-100} & \multicolumn{2}{c}{\bf Split ImageNet-R}\\ 
         & Avg. Acc ($\uparrow$) & Forget ($\downarrow$)
        & Avg. Acc ($\uparrow$) & Forget ($\downarrow$)\\
        \hline 
        
        LoRAC w/o TII & 90.80 &6.04 &78.29 &7.07\\
        +Random  & 90.65 & 5.86 &78.45 &6.31\\  
        \CC{ours}+IPC & \CC{ours}\bf 91.25 & \CC{ours}\bf 5.34 &\CC{ours}\bf 78.85 &\CC{ours}\bf 4.09\\
    \bottomrule
\end{tabular}}
\vspace{-10pt}
\end{table}

\noindent \textbf{Effect of important parameters constrain (IPC).} 
In \cref{sec:ipc}, we have defined the average importance of the parameter matrix across different blocks in the ViT.
To validate the effectiveness of our method for estimating importance, ensuring that model parameters crucial for the current task can indeed be identified by our approach, we performed the following experiments. 
We randomly select some parameter matrices as the important parameters for the current task, referred to as \textbf{Random}, for comparison with IPC.
As can be seen from the results in \Cref{table:ipc}, IPC outperforms Random on both datasets, proving that IPC is effective at identifying model parameters critical to the current task and mitigates model's forgetting by freezing those parameters in subsequent tasks.

\begin{table}[ht]
\centering
\caption{Ablation study of hierarchical components on Split CIFAR-100.}
\label{table:ablation study 1}
\resizebox{0.48\textwidth}{!}{
\begin{tabular}{llcc}
    \toprule
        \multirow{2}{*}{\bf PTM} &\multirow{2}{*}{\bf Method}   &\multicolumn{2}{c}{\bf Split CIFAR-100}\\
        & & Avg. Acc ($\uparrow$) & Forget ($\downarrow$)\\
        \hline 
        \multirow{5}{*}{\bf Sup-21K*}
        &LoRA-FT  & 84.41 &14.74 \\
        &+Composition  & 86.37&12.62 \\
        &+Orthogonal Loss  & 90.80 & 6.04 \\  
        &+IPC & 91.25 & 5.34 \\
        &+TII  & \bf{92.14} & \bf{2.49} \\
    \bottomrule
\end{tabular}}
\vspace{-7pt}
\end{table}

\noindent \textbf{Effect of key components for \FrameworkNameIPC{}.}  \Cref{table:ablation study 1} presents a systematic analysis of the incremental incorporation of the key components of our method: LoRA composition, orthogonality regularization, important parameters contraints(IPC) and task ID inference (TII).  
As shown in \Cref{table:ablation study 1}, adding composition improves accuracy by $1.96\%$, while forgetting decreases by $2.12\%$. 
We analyze this performance improvement by combining the previous task-specific LoRA weights with weights coefficients to retain knowledge from previous tasks. 
Furthermore, our findings reveal that incorporating orthogonal constraints into the LoRA composition yields significant performance enhancement: accuracy improves by an additional $4.43\%$, while the forgetting decreases by $6.58\%$. 
This indicates that imposing orthogonality constraints on the model effectively mitigates interference between prior and new knowledge, enhancing the adaptability of the model.
In addition, when combined with IPC, which prevents the model from changing parameters important to previous tasks when learning new tasks, thereby reducing interference with previous knowledge, the average accuracy of the model is further improved by $0.45\%$, and the forgetting rate decreases by $0.7\%$.
Finally, task ID inference resulted in an additional 0.89\% increase in accuracy and a 2.85\% decrease in forgetting.

\begin{table}[ht!]
\centering
\caption{Results for rehearsal-free continual learning on UESTC-MMEA-CL.}
\label{table:Multimodel}
\resizebox{0.48\textwidth}{!}{
\begin{tabular}{lcccc}
    \toprule
    \multirow{2}{*}{\bf Method} &\multicolumn{2}{c}{\bf 4 Tasks} &\multicolumn{2}{c}{\bf 8 Tasks} \\
        & Avg. Acc ($\uparrow$) & Forget ($\downarrow$) & Avg. Acc ($\uparrow$) & Forget ($\downarrow$) \\
        \midrule
        LoRA-FT &83.87 &12.63 &73.36 &19.33 \\
        \CC{ours}\FrameworkNameIPCWo{} &\CC{ours}94.15 &\CC{ours}5.26 &\CC{ours}87.77 &\CC{ours}8.21 \\
        \CC{ours}\FrameworkNameIPC{} &\CC{ours}\textbf{95.04} &\CC{ours}\textbf{1.21} &\CC{ours}\bf{92.96} &\CC{ours}\bf{2.52} \\
    \bottomrule
    \label{tab:mmcl}
\end{tabular}}
\vspace{-25pt}
\end{table}

\section{Discussion on Multi-Modal Continual Learning}

We further validate the effectiveness of LoRAC-IPC on the multimodal continual learning dataset UESTC-MMEA-CL~\citep{xu2023towards}. Some examples are shown in Appendix.~B. UESTC-MMEA-CL contains three data modalities: video data, accelerometer data, and gyroscope data, consisting of 32 classes of daily activities. For the video data, we uniformly sample 8 frames from each video and use ViT-B/16 to extract features from each frame, then average the features of the 8 frames. The model is initialized with Sup-21K*. For the accelerometer and gyroscope data, we use STFT to convert them into spectrograms and then use Tiny-SSAST to extract features from the spectrograms, initializing the model with weights pre-trained on AudioSet and Librispeech. We then concatenate the features from the three types of data and use a linear head to predict the classes. We conduct experiments on two standard settings of UESTC-MMEA-CL: 4 tasks and 8 tasks splitting. The experimental results are shown in Tab.~\ref{tab:mmcl}. It can be observed that LoRAC-IPC achieves better accuracy and forgetting compared to LoRA-FT in both 4 tasks and 8 tasks splitting, especially in the longer 8 tasks splitting, where LoRAC-IPC's performance is more significant, with a $19.6\%$ higher accuracy than LoRA-FT. It is worth noting that after using TII, the performance of LoRAC-IPC has improved further.

\section{Conclusion}
\label{sec:Con}


We propose LoRAC-IPC, a LoRA composition-based continual learning method with constrains on critical parameter changes. On the one hand, LoRA composition preserves old knowledge while introducing new knowledge by integrating pre-trained model parameters with orthogonal LoRA modules. 
And the learnable weights on LoRA modules promote the plasticity of our method.
On the other hand, important parameter constraints (IPC) force the critical parameters of the current task to remain unchanged during subsequent learning, leading to further reduced forgetting.
Extensive experimental results comprehensively demonstrate the effectiveness of LoRAC-IPC.

Some further studies are left in the future. Firstly, the computational complexity of task ID inference of the proposed method is costly. Optimizing the task ID inference to reduce the computational complexity is a future concern. Secondly, the performance of the proposed method and potential improvements can be explored in more general and challenging settings in continual learning. We hope our work may inspire further study of visual continual learning combined with pre-trained models.



\section*{Acknowledgement}
This work is supported by the National Key R\&D Program of China (2021ZD0112001) and the National Natural Science Foundation of China (No.~62171111).

\appendix

\section{Implementation Details}
\begin{table*}[hbtp]
\vspace{-7pt}
\begin{center}
    \caption{The details about the weight type and rank of \FrameworkName{} and \FrameworkNameIPC{}}
    \label{tab:sup1}
    \resizebox{1.0\linewidth}{!}{
    \begin{tabular}{l l cc cc cc cc}
    \toprule
        \multirow{2}{*}{\textbf{PTM}} &\multirow{2}{*}{\textbf{Mehods}}& \multicolumn{2}{c}{\textbf{Split CIFAR-100}} & \multicolumn{2}{c}{\textbf{5-datasets}}& \multicolumn{2}{c}{\textbf{Split ImageNet-R}} &\multicolumn{2}{c}{\textbf{Split DomainNet}}\\
        & &Weight Type&Rank &Weight Type&Rank&Weight Type&Rank &Weight Type&Rank\\
        \midrule
        \multirow{1}*{Sup-21K}
        &\FrameworkName{} / \FrameworkNameIPC{}  &$\mathbf{W}_{all}$ &32 & $\mathbf{W}_{K}$, $\mathbf{W}_{V}$ & 8 &$\mathbf{W}_{all}$ &64 &-&-\\
        \midrule
        \multirow{1}*{Sup-21K*}
        &\FrameworkName{} / \FrameworkNameIPC{}&$\mathbf{W}_{all}$ & 16 & - & - & $\mathbf{W}_{all}$& 32 &$\mathbf{W}_{Q},\mathbf{W}_{K},\mathbf{W}_{V}, \mathbf{W}_{O}$&64\\
        \midrule
        \multirow{1}*{MoCo-1K}
        &\FrameworkName{} / \FrameworkNameIPC{} &$\mathbf{W}_{all}$ &16 &- &- &$\mathbf{W}_{all}$ &16 &-&-\\
    \bottomrule
    \end{tabular}
    }
\end{center}
\vspace{-10pt}

\end{table*}

\begin{figure*}[ht!]
    \centering
    \subfloat[sit-stand]{\includegraphics[width=0.45\textwidth]{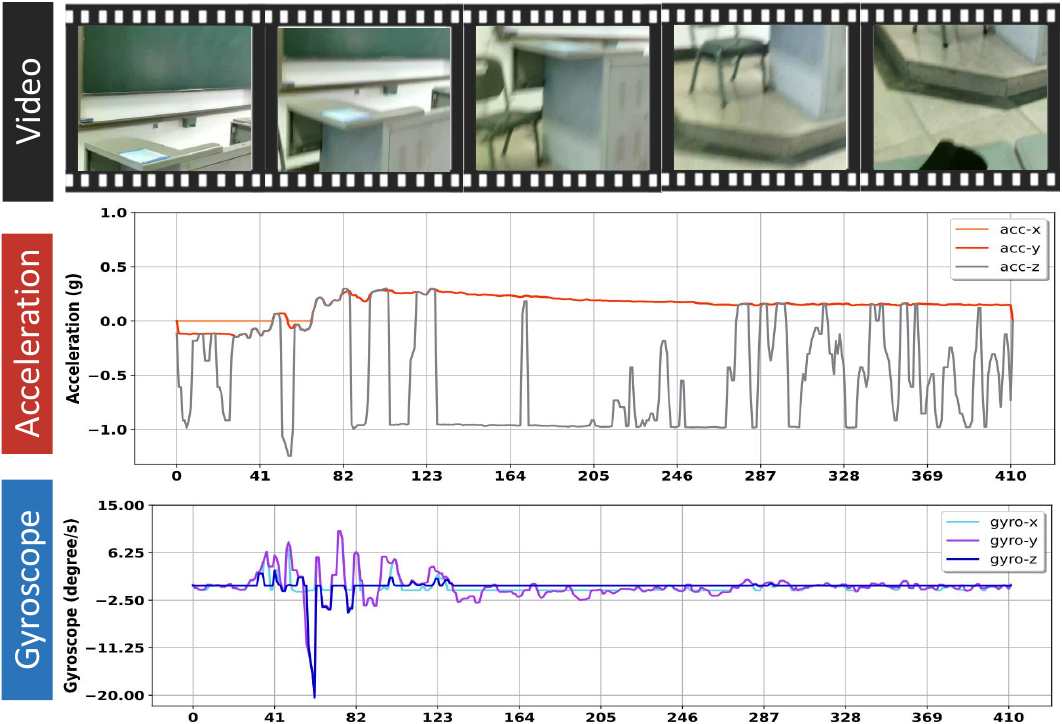}
    }
    \subfloat[downstairs]{\includegraphics[width=0.45\textwidth]{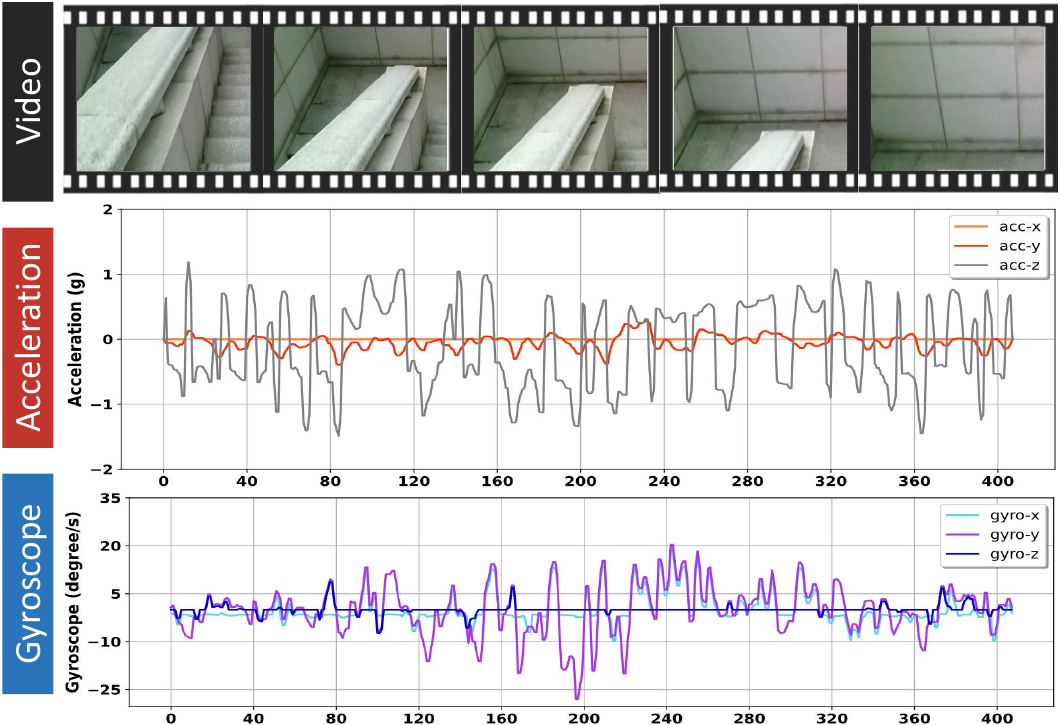}
    }\\
    
    \subfloat[walking]{\includegraphics[width=0.45\textwidth]{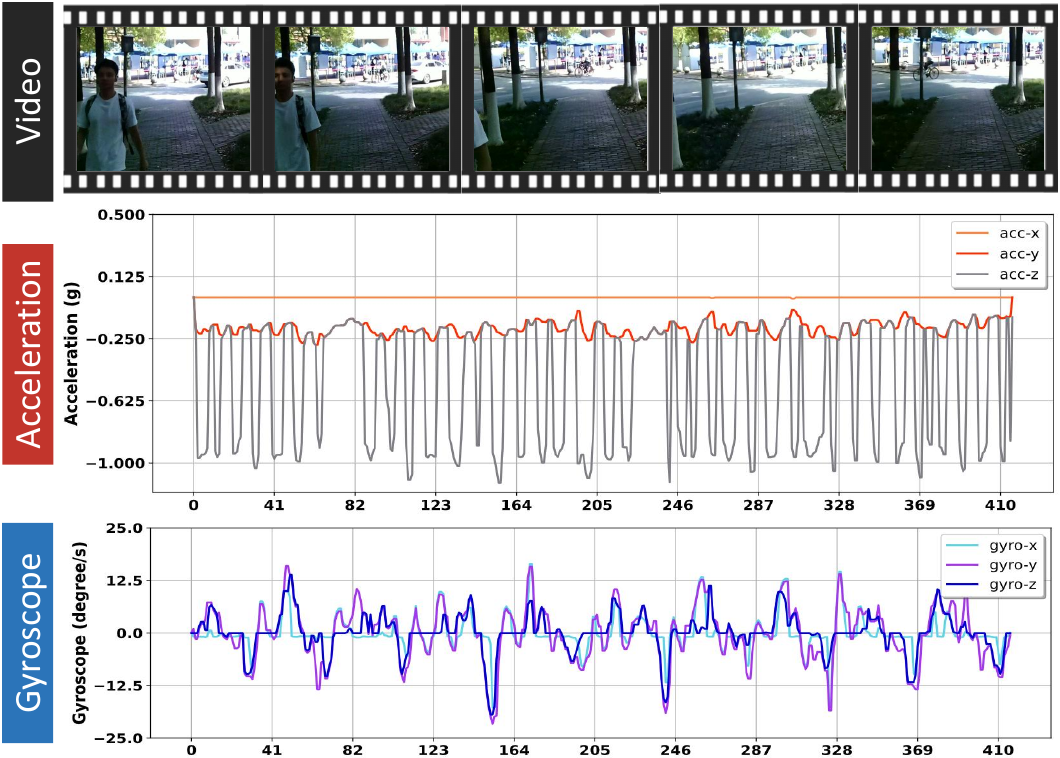}
    }
    \subfloat[play-phone]{\includegraphics[width=0.45\textwidth]{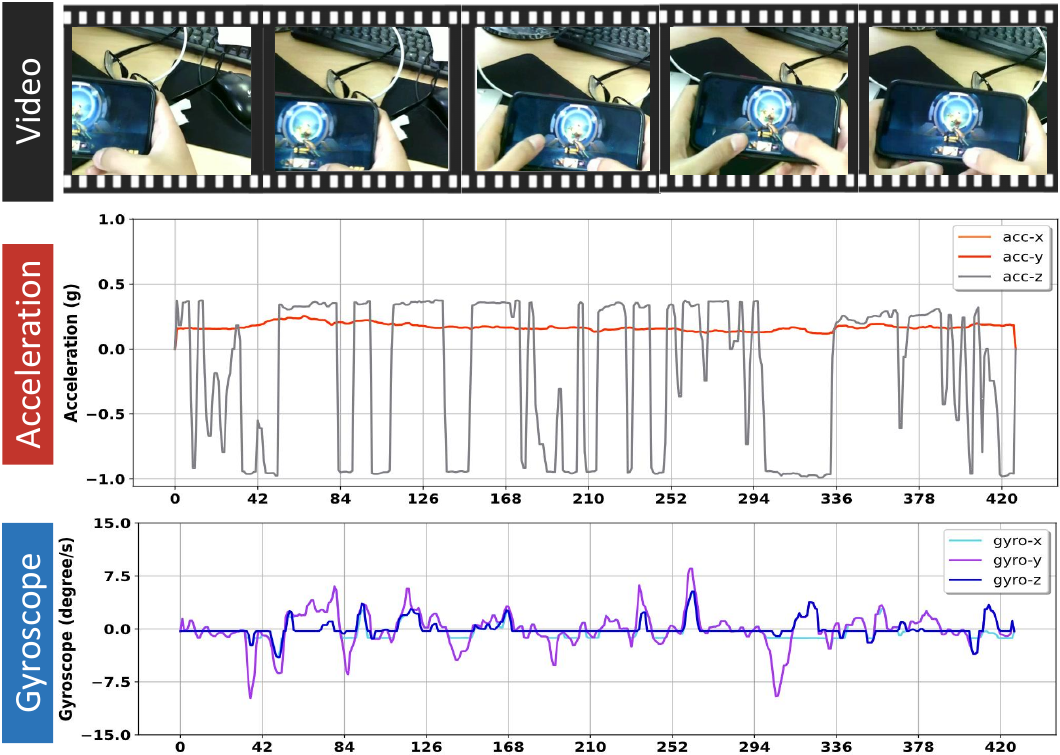}
    }\\

    \caption{Representative examples from UESTC-MMEA-CL}
    \label{fig:uestc}
    \vspace{-10pt}
\end{figure*}

Following the implementations of previous work\citep{wang2023hide, wang2022learning}, we employ a pre-trained ViT-B/16 backbone, an Adam optimizer and a batch size of $128$. 
The $\mathbf{A}_{\tau}(\tau=1,2,...T)$ matrix is initialized using the Kaiming initialization, and the $\mathbf{B}_{\tau} (\tau=1,2,...T)$ is initialized to zero. 
The weight coefficients are initialized to 1 and subsequently updated with a small learning rate.

\noindent \textbf{Sup-21K}. 
The configuration for our experiments on Split CIFAR-100, 5-datasets, and Split ImageNet-R is as follows. Firstly, training durations vary, with the model undergoing 20 epochs for Split CIFAR-100, 20 epochs for 5-datasets, and 50 epochs for Split ImageNet-R. Secondly, the trade-off values($\lambda$) of the orthogonal loss also vary across the three datasets:  Split CIFAR-100 ($\lambda=1.0$), 5-datasets ($\lambda=1e-6$) and Split ImageNet-R ($\lambda=0.01$).

\noindent \textbf{Sup-21K*}. The training durations and orthogonal loss trade-off values ($\lambda$) are set as follows: for Split CIFAR100, 10 epochs and 0.1; for Split ImageNet-R, 50 epochs and 1.0; and for Split DomainNet, 50 epochs and 1.0.

\noindent \textbf{MoCo-1K}. The training duration is set to 20 epochs for each dataset. For Split CIFAR-100, the orthogonal loss trade-off value ($\lambda$) is set at 0.1, while for Split ImageNet-R, it is also set at 0.1.

\subsection{Types of weights adapted with LoRAC.}
In this section, we report the modules of the ViT model where \FrameworkName{}/\FrameworkNameIPC{} will be added for different pre-trained models. \Cref{tab:sup1} shows the types of weights adapted with \FrameworkName{}/\FrameworkNameIPC{}, where $\mathbf{W}_{Q}$, $\mathbf{W}_{K}$, and $\mathbf{W}_{V}$ represent the parameters of the QKV projection matrices of the multi-head self-attention blocks, while $\mathbf{W}_{O}$ denotes the parameters of the subsequent linear mapping layer. $\mathbf{W}_{fc1}$ and $\mathbf{W}_{fc2}$ represent the parameters of the first and second linear layers in the MLP block, respectively. $\mathbf{W}_{all}$ denotes that LoRAC is added to all modules ($\mathbf{W}_{Q}, \mathbf{W}_{K}, \mathbf{W}_{V}, \mathbf{W}_{O},\mathbf{W}_{fc1}, \mathbf{W}_{fc2}$) throughout the ViT model.

\section{More Details and Results on Multi-model Continual learning}

\noindent \textbf{Examples of UESTC-MMEA-CL.} UESTC-MMEA-CL\citep{xu2023towards} is a multi-modal first-person activity dataset for continuous egocentric activity recognition. 
\Cref{fig:uestc} shows examples of UESTC-MMEA-CL datasets.


\noindent \textbf{More Experimental Results on Multi-model Data.} We further conduct experiments on the multi-modal dataset ARIC~\citep{aric} to verify the effectiveness of LoRAC-IPC. The ARIC dataset, which is derived from real classroom surveillance, encompasses 32 classroom activities across three modalities: image, text, and audio. The dataset is divided into 4 tasks, each containing 8 classes. The experimental results are presented in Tab.~\ref{tab:aric}. It can be observed that LoRAC-IPC outperforms LoRA-FT in terms of both accuracy and forgetting.
\begin{table}[ht!]
\centering
\caption{Results for rehearsal-free continual learning on ARIC.}
\resizebox{0.35\textwidth}{!}{
\begin{tabular}{lcc}
    \toprule
    \multirow{2.5}{*}{\bf Method} &\multicolumn{2}{c}{\bf ARIC} \\
        \cmidrule(lr){2-3}
        & Avg. Acc ($\uparrow$) & Forget ($\downarrow$) \\
        \midrule
        LoRA-FT &40.53 &28.78  \\
        \CC{ours}\FrameworkNameIPCWo{} &\CC{ours}57.70 &\CC{ours}23.04 \\
        \CC{ours}\FrameworkNameIPC{} &\CC{ours}\textbf{67.08} &\CC{ours}\textbf{2.56} \\
    \bottomrule
    \label{tab:aric}
\end{tabular}}
\vspace{-25pt}
\end{table}

\bibliographystyle{elsarticle-num}
\bibliography{refs}





\end{document}